\pdfoutput=1
\documentclass[11pt]{article}
\usepackage[table]{xcolor}
\usepackage[final]{acl}

\usepackage{hyperref}
\usepackage{url}
\usepackage{multirow}

\usepackage{booktabs}
\usepackage{multicol}
\usepackage{graphicx}
\usepackage{bbm, dsfont}
\usepackage{makecell} % Add this package

\usepackage{wrapfig}
\usepackage{subcaption}

\usepackage{float}
\usepackage{amsmath}

\usepackage{times}
\usepackage{latexsym}

\usepackage[T1]{fontenc}
\usepackage[utf8]{inputenc}

\usepackage{microtype}

\usepackage{inconsolata}

\usepackage{graphicx}

\title{GraphNarrator: Generating Textual Explanations for Graph Neural Networks}

\author{%
    Bo Pan$^{*}$, Zhen Xiong$^{*}$, Guanchen Wu$^{*}$, Zheng Zhang, Yuntong Hu, Yifei Zhang,\\ \textbf{Liang Zhao$^{\dagger}$} \\
    Emory University, Atlanta, USA \\ 
    \texttt{\{bo.pan, guanchen.wu, liang.zhao\}@emory.edu, xiongzhen0711@gmail.com}
\\
}

\begin{document}

\maketitle

\renewcommand{\thefootnote}{\fnsymbol{footnote}}  % Use symbols for footnotes
\footnotetext[1]{These authors contributed equally.}
\footnotetext[2]{Corresponding author.}

\begin{abstract}
Graph representation learning has garnered significant attention due to its broad applications in various domains, such as recommendation systems and social network analysis. Despite advancements in graph learning methods, challenges still remain in explainability when graphs are associated with semantic features. In this paper, we present GraphNarrator, the first method designed to generate natural language explanations for Graph Neural Networks. GraphNarrator employs a generative language model that maps input-output pairs to explanations reflecting the model's decision-making process. To address the lack of ground truth explanations to train the model, we propose first generating pseudo-labels that capture the model's decisions from saliency-based explanations, then using Expert Iteration to iteratively train the pseudo-label generator based on training objectives on explanation quality. The high-quality pseudo-labels are finally utilized to train an end-to-end explanation generator model. Extensive experiments are conducted to demonstrate the effectiveness of GraphNarrator in producing faithful, concise, and human-preferred natural language explanations.

\end{abstract}

\begin{figure}[t]
    \centering
    \includegraphics[width=1\linewidth]{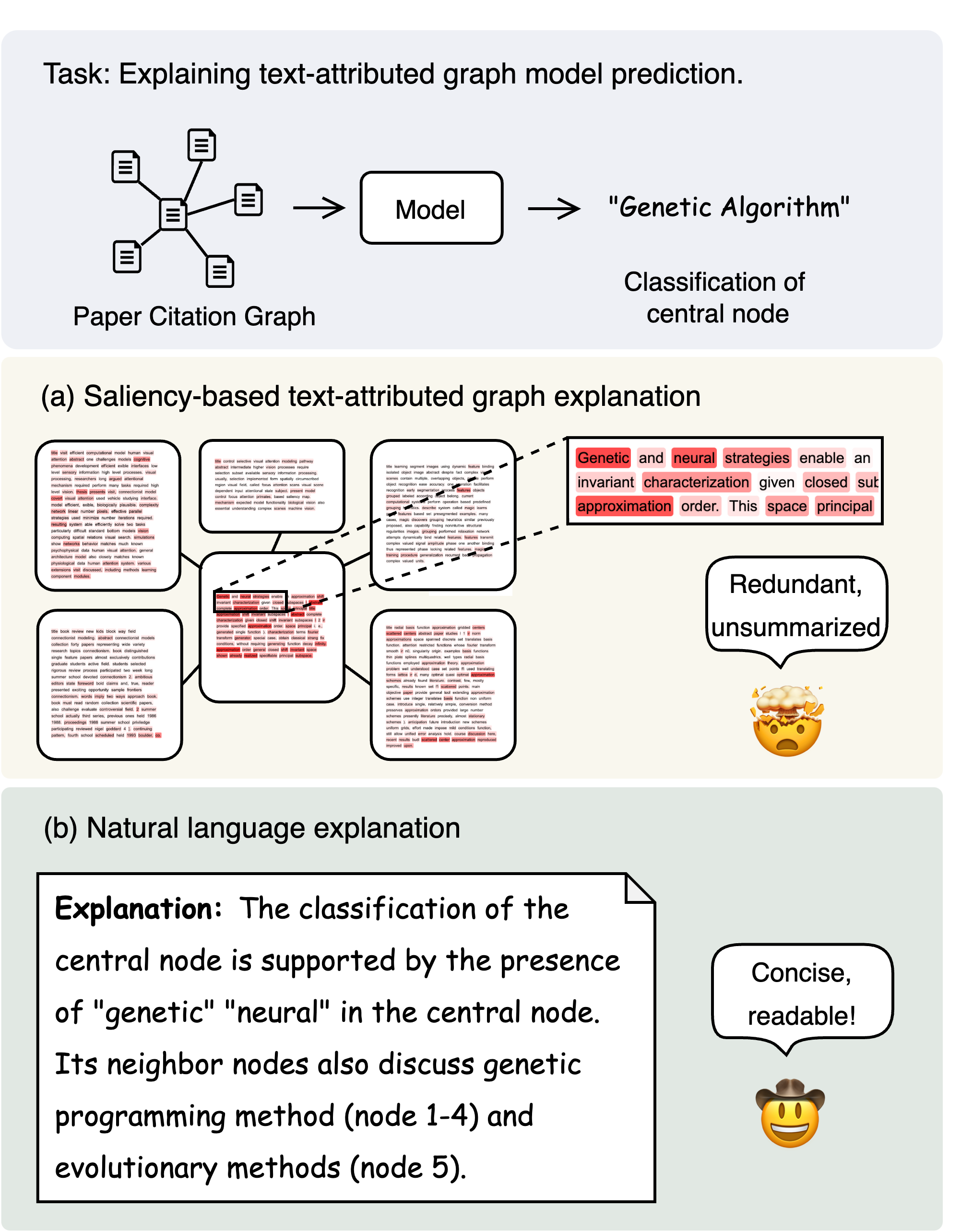}
    \caption{Illustration of saliency-based graph explanation and natural language graph explanation.}
    \label{fig:intro}
    \vspace{-5mm}
\end{figure}

\section{Introduction}
Deep learning on graphs, especially Graph Neural Networks (GNNs) \citep{kipf2016semi, hamilton2017inductive}, has become the standard and dominant technique for various graph-related tasks due to its expressive power in capturing structured features useful for prediction and representation learning. In many real-world scenarios, graph nodes and edges are accompanied by textual features, either inherently textual as in Text-Attributed Graphs (TAGs) \cite{zhou2019gear, yang2021graphformers} or described using semantic terms and numerical expressions.  For example, in e-commerce graphs of recommendation systems, products can be associated with textual descriptions; in molecular graphs, atoms and bonds can be described by textual tokens such as element names or properties. These textual descriptions offer rich semantic cues that can augment graph learning \cite{he2023harnessing, pan2024distilling, hu2024grag}. Despite the dominant performance of current GNNs across a range of tasks, their internal decision processes remain largely opaque, which hampers their adoption in high-stakes applications. In this work, we aim to explore the right explanation modality for graphs with revolutionary tradeoff between conciseness, fidelity, and readability.

% Although current LLMs can generate rationales in a zero-shot manner, these rationales cannot faithfully reflect the decision-making process since the LLMs have no access to the inner decision process of models \citep{agarwal2024faithfulness, parcalabescu2024measuring}. In this work, we dive into the problem of explaining TAG model predictions, which is an important yet not well-explored area. 

Explainability for graph learning have received a large amount of attention, with methods usually by providing node- or edge-level importance scores \citep{ying2019gnnexplainer,vu2020pgm,luo2020parameterized}, but they lack the ability to explain the semantic information in graphs since the node- and edge-level importance scores cannot include any information of the text features. Some methods \citep{ying2019gnnexplainer, vstrumbelj2014explaining, bach2015pixel} can also provide node feature-level explanation, and thus can generate token importance scores for textual features on nodes. However, graph predictions are often made based on a subgraph with many nodes and their associated texts, such token importance-based explanations are hardly human-understandable as they are redundant and not integrated, as illustrated in Fig.~\ref{fig:intro}~(a). The limitation and need for human understandability necessitates our research on natural language explanations for GNNs, which is expected to be summative, concise, and readable, as one example shown in Fig.~\ref{fig:intro}~(b).

In this work, we present GraphNarrator, the first method to generate natural language explanations for GNNs. GraphNarrator is a model-agnostic method to learn a mapping from graph model input-output pairs to textual explanations. To generate textual explanations, since external language models have no knowledge of the model's internal decision-making process to generate high-quality explanations, it is necessary to fine-tune the model with high-quality explanation labels. In real-world scenarios, it is impractical to have annotated ground truth data for explaining model behaviors, therefore we propose graph explanation verbalization to prompt LLMs with saliency-based explanations to generate natural language explanation pseudo-labels. To continuously improve the quality of pseudo-labels, we propose three training objectives related to faithfulness and brevity, and iteratively fine-tune the pseudo-label generator model with these objectives with expert iteration. Finally, the generated pseudo-labels are used to train an end-to-end explainer model, which serves as our end-to-end explanation generation model. 

% Our contribution can be summarized as follows: 
% \begin{itemize}
%     \item We propose GraphNarrator, a novel framework to generate natural language explanations for TAG learning models. GraphNarrator is a model-agnostic black-box model explanation method that can generate post-hoc explanations based on the input and prediction.
%     \item We propose to narrate saliency-based TAG explanations by our TAG explanation verbalization and prompting. Our graph verbalization preserves the semantic information, structural information as well as feature importance, and forms a more LLM-understandable format of input to be prompted the explanation pseudo-label generator LLM.
%     \item We propose to iteratively self-train the explanation pseudo-label generator LLM via TAG Explanation Expert Iteration with three objectives for improving the graph explanation quality, considering the faithfulness to important input, faithfulness to output, and brevity.
%     \item We conducted extensive experiments to validate the effectiveness of our proposed framework. Experiments show that GraphNarrator can self-improve for label generation and generate more faithful and brief explanations.
% \end{itemize}

\begin{figure*}[ht!]
    \centering
    \vspace{-3mm}
    \hspace{2mm}
    \includegraphics[width=0.95\linewidth, page=1]{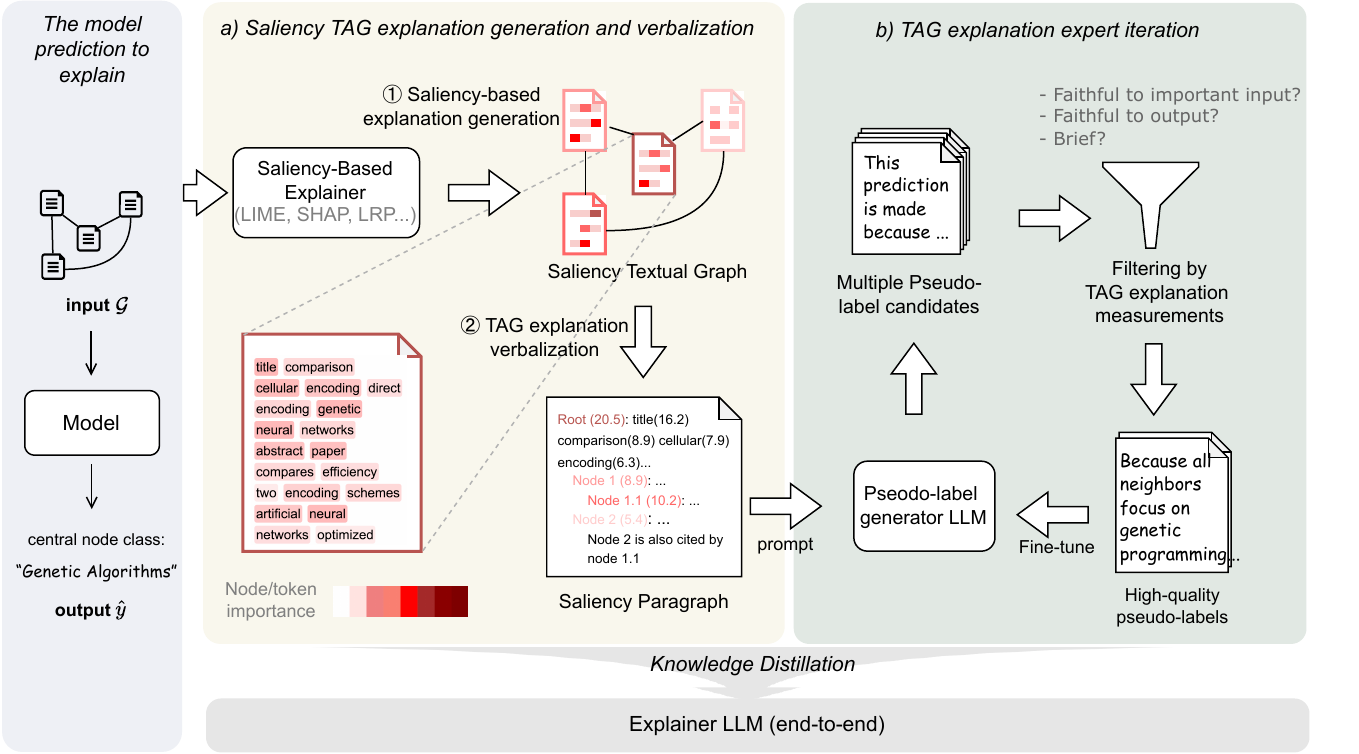}
    \caption{An illustration of GraphNarrator. A pseudo-label generator model is first trained to provide pseudo-labels, which are used for knowledge distillation to an LLM as an end-to-end explainer. (a) GraphNarrator first generates saliency-based graph explanations, then verbalizes them into a documented form (\textit{Saliency Paragraph}) for easier understanding of LLMs, and feeds them to LLMs to generate initial natural language explanation pseudo-labels. (b) We propose the graph explanation expert iteration procedure to iteratively improve the pseudo-label generator LLM with three objectives related to faithfulness and brevity. }
    \label{fig:method}
    \vspace{-4mm}
\end{figure*}

\section{Related Work}
\subsection{Explainability of Graph Neural Networks} 

GNNs have been widely adopted in fields such as social networks, molecular chemistry, and financial systems, yet their interpretability remains a significant challenge. Existing GNN explanation methods are generally categorized into instance-level and model-level approaches. Model-level methods seek to provide a broad understanding of GNN decision behaviors independent of specific inputs~\citep{zhang2022protgnn, yuan2020xgnn}, while instance-level methods focus on explaining individual predictions by identifying important features that influence the model’s decisions. These methods include gradients/features-based approaches~\citep{baldassarre2019explainability}, perturbation-based methods~\citep{ying2019gnnexplainer, luo2020parameterized}, decomposition-based techniques~\citep{baldassarre2019explainability,schwarzenberg2019layerwise} and surrogate model-based methods~\citep{huang2022graphlime}. 
While significant progress has been made in improving the explainability of GNNs, current instance-level methods are limited to providing feature importance-based explanations, which is difficult for humans to understand when the input comes to TAGs. To our best knowledge, no efforts have been made to generate natural language explanations of graph models. 

\subsection{Natural Language Explanation}
The traditional explanation methods for NLP, such as feature importance and saliency maps~\citep{lei2016rationalizing, yu2019rethinking}, often fall short in providing human-interpretable insights, motivating the development of more intuitive approaches like natural language explanations~\citep{cambria2023survey}. These methods provide textual justifications for model predictions, aiming to bridge the gap between model decisions and human understanding~\citep{camburu2018snli,rajani2019explain,narang2020wt5}. A notable example is self-explanation models, where models predict labels while simultaneously generating explanations in natural language ~\citep{wiegreffe2020measuring, zhang2024elad, liu2023d, liu2024enhancing}. With the advent of LLMs, Chain-of-Thought prompting~\citep{wei2022chain} and zero-shot reasoning~\citep{kojima2022large} have enhanced their self-explanation abilities by generating coherent, step-by-step explanations. Additionally, LLMs’ role as explainers for both their own predictions and other models’ outputs has significantly expanded~\citep{kroeger2023large, gat2023faithful, martens2023tell}. Recent research also points out that LLM-based explanations do not necessarily faithfully reflect the internal model behaviours \citep{agarwal2024faithfulness, parcalabescu2024measuring}.

\section{Problem Formulation} 
In this work, we delve into the task of explaining predictions for GNNs with natural languages. Formally, a Text-Attributed Graph (TAG) can be represented as $\mathcal{G}=\left(\mathcal{V}, A,\mathcal{X}\right)$, where $\mathcal{V}=\{v_0, v_1, ..., v_{N-1}\}$ is a set of $N$ nodes, $A \in \{0,1\}^{N \times N}$ is the adjacency matrix, and $\mathcal{X}=\{x_0, x_1, ..., x_{N-1}\}$ is the set of texts where $x_k=(t_{k,0}, t_{k,1}, ..., t_{k,n^{(t)}_k})$ is a sequence of tokens associated with node $v_k \in \mathcal{V}$. 
A TAG model $f$ is a model that can make predictions on TAGs by $\hat y=f(\mathcal{G})$, where $\hat y$ is the model output. 

Given a text-attributed graph $\mathcal{G}$ and a trained TAG model $f$, the goal is to learn a mapping $g: (\mathcal{G}, \hat y)\to E$ to generate a paragraph of text $E$ to explain the decision-making process of $\hat y=f(\mathcal{G})$.
The generated explanation $E$ should faithfully explain the reason for model predictions and be friendly for humans to understand.

\section{Proposed Method: GraphNarrator}

The overall framework of GraphNarrator is illustrated in Figure~\ref{fig:method}. The high-level idea is we first train a \textit{Pseudo-Label Generator LLM} to generate high-quality explanation labels, then the generated pseudo-labels are used to fine-tune an \textit{Explainer LLM}, which serves as a post-hoc explainer, the product of our framework. Due to the lack of ground truth explanation labels, we first use a saliency-based explainer to attain \textit{Saliency Textual Graph} explanation, then verbalize it to the form of a \textit{Saliency Paragraph}, and use it as a hint of the important regions for the \textit{Pseudo-Label Generator LLM} (Sec.~\ref{sec:verbalize}). Then the \textit{Pseudo-Label Generator LLM} iteratively self-improves via Expert Iteration based on our proposed three training objectives for quantifying explanation quality (Sec.~\ref{sec:expert}). Finally, the generated high-quality pseudo-labels from the fine-tuned \textit{Pseudo-Label Generator LLM} are used to train the end-to-end \textit{Explainer LLM} (Sec.~\ref{sec:distill}). The details of our proposed GraphNarrator framework will be introduced in the following.

\subsection{Saliency-based Graph Explanation Generation and Verbalization}\label{sec:verbalize}
We first construct a \textit{Saliency Paragraph} to pass the semantics and structural information of TAG data as well as the saliency information of the model decision, as a hint for the \textit{Pseudo-Label Generator LLM}. As illustrated in Fig.~\ref{fig:method}~(a), specifically, a saliency-based explainer is first adopted to get the \textit{Saliency Textual Graph} explanations, and then we verbalize the graph explanation into textual forms based on BFS and hierarchical organizing. Finally, we prompt them to the \textit{pseudo-label generator LLM} to get the initial explanation of pseudo-label candidates. Such a procedure will be introduced in details as follows.

\textbf{Saliency-based explanation generation.} 
As shown in Fig.~\ref{fig:method}~(a), the saliency-based explanations are generated by a feature importance-based post-hoc explainer. They are represented in the form of the importance score of each node and token. An example is illustrated in Fig.~\ref{fig:method}~(a) as \textit{Saliency Textual Graph}, where the red color from light to dark denotes the importance of nodes and tokens.
% Formally, for a model prediction $y=f(\mathcal{G})$ for the input graph $\mathcal{G}=\left(\mathcal{V}, A,\mathcal{X}\right)$, the saliency-based explanation $X=(X^{(\mathcal N)}, X^{(\mathcal E)}, X^{(\mathcal T)})$, where $X_N$, $X_E$ and $X_T$ denoted the corresponding importance scores for nodes, edges and tokens towards the prediction of label $y$, and $X^{(\mathcal N)}=\{s_k^{(\mathcal N)}\}_{k=0}^{N-1}$, where $s_k^{(\mathcal N)}$ is the importance of node $v_k$; $X^{(\mathcal T)}=\{s_0^{(\mathcal T)}, s_1^{(\mathcal T)}, ..., s_{N-1}^{(\mathcal N)}\}$, where $s_k^{(\mathcal N)}$ is the importance scores of all tokens in $v_k$, and $s_k^{(\mathcal N)}=(s_{k,0}^{(\mathcal N)}, s_{k,1}^{(\mathcal N)},... , s_{k,n^t_k-1}^{(\mathcal N)})$. 
Note that here the saliency-based explainer can be based on various methods that can generate post-hoc explanations for a TAG model, including widely used model-agnostic explanation methods including LRP, Input$\times$Grad, Saliency, etc. This makes GraphNarrator a model-agnostic framework for explaining various TAG learning model architectures. 

\textbf{Saliency textual graph verbalization.} Since the generated \textit{Saliency Textual Graph} is graph-structured data, it is vital to transform it into a form that LLMs are easier to understand. Therefore, we propose saliency textual graph verbalization, to transform the saliency-based graphs explanation into a document-like \textit{Saliency Paragraph} to pass the structural, semantic and feature importance information to LLMs.

In a TAG model prediction, the structure of a central node and its k-hop salient nodes can be represented as an ego graph, with the node itself as the root. Using Breadth-First Search (BFS), this ego graph can be decomposed into a tree structure. During the BFS process, we prune unimportant nodes by identifying nodes with no tokens whose saliency scores are higher than a threshold. Then we adopt a Pre-Order Traversal (first visit the root, then the left subtree, then the right subtree) to organize the tree structure into a hierarchical saliency paragraph, maintaining the hierarchical structure of the k-hop ego graph. In this Saliency Paragraph, each node's text is represented as a section, and its successor nodes are represented as subsections. Note that when converting the ego graph to a BFS tree, there can be a set of \textit{cross-edges} that connect nodes in different branches. We verbalize them by adding reference sentences in the text of source nodes, pointing to the sections representing their respective destination nodes. This ensures that the saliency paragraph faithfully reflects the ego graph's structure.

\begin{figure}[th]
    \centering
    \vspace{-2mm}
    \includegraphics[width=1\linewidth]{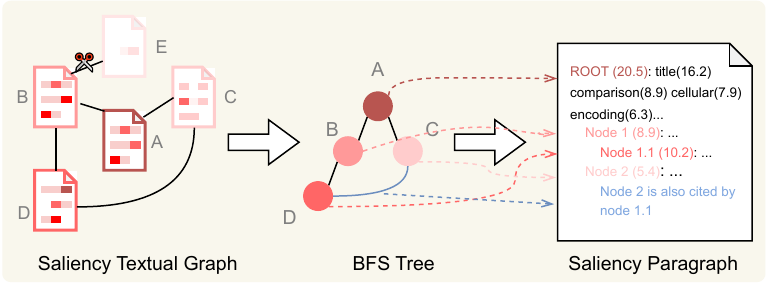}
    \vspace{-7mm}
    \caption{Illustration of graph explanation verbalization. The blue edge denotes a cross-edge.}
    \vspace{-2mm}
    \label{fig:bfs}
\end{figure}

When organizing the \textit{Saliency Paragraph}, we attach the saliency scores of nodes and tokens to prompt the \textit{Pseudo-Label Generator LLM} with the feature importance information. For tokens in nodes, we attach their importance score after the tokens to prompt with the importance information, in the form of \verb|token(score)|, without perturbing the semantic order of tokens. An example of constructing the \textit{Saliency Paragraph} is given in Figure~\ref{fig:bfs}.

\subsection{TAG Explanation Expert Iteration for Explanation Self-Improvement}\label{sec:expert}

Initial explanations can be attained by simply passing the generated \textit{Saliency Paragraph} to an LLM to generate more summarized words. However, such a zero-shot generation manner cannot ensure the explanation quality and may still lead to unfaithful or redundant textual explanations. To quantify the explanation quality, we propose Information-Theoretic TAG Explanation Measurements that quantify the text explanations' faithfulness and brevity. To effectively optimize such goals without ground truth labels, we propose TAG Explanation Expert Iteration framework that iteratively conducts \textit{text explanation quality measuring}, \textit{high-quality text explanation selecting}, and \textit{text explanation updating}, as illustrated in Fig.~\ref{fig:method}~(b). 

\subsubsection{Information-Theoretic TAG Explanation Measurements}\label{sec:objectives}
High quality model explanations should be faithful to the model decision process, and friendly for humans to understand. Therefore, we propose our TAG explanation measurements, quantify the explanation quality with faithfulness to important inputs, faithfulness to outputs, and brevity.

\textbf{Faithfulness to important inputs.} The explanation is expected to include enough necessary information about the rationale of the model prediction. Suppose there is a true rationale $\mathcal R$, then the faithfulness can be measured by the Pointwise Mutual Information (PMI) between the generated explanation $E$ and $\mathcal{R}$, namely,

\begin{equation}
    % f_S=\text{PMI}(E, \mathcal{R})= \log \frac{P(\mathcal{R}, E)}{P(\mathcal{R})P(E)} = \log \frac{P(\mathcal{R}| E)}{P(\mathcal{R})} 
    f_S=\text{PMI}(E, \mathcal{R})= \log \frac{P(\mathcal{R}| E)}{P(\mathcal{R})} 
\end{equation}
\vspace{-2mm}

However, estimating $P(\mathcal R)$ and $P(\mathcal R|E)$ is intractable due to the big space of $\mathcal R$ to explore. Therefore, we formulate $\mathcal R$ as the subset of important nodes and tokens in the n the input graph $\mathcal G$ (the serialization of $\mathcal G$ can be found in Appendix~\ref{append:serialize}). We propose to mask the important tokens in nodes of $\mathcal G$, and turn this problem into a tractable masked token prediction problem with a pretrained language model. Therefore, $P(\mathcal{R})$ and $P(\mathcal{R}|E)$ can be estimated with $P(\mathcal{R}|\mathcal G_M)$ and $P(\mathcal{R}|\mathcal G_M, E)$, where $\mathcal G_M$ denotes the remaining part in $S$ after masking $\mathcal{R}$ from $\mathcal G$. Moreover, we also need to consider the fact that not all the tokens in $\mathcal R$ are equally important, so we want to prioritize the faithfulness to those most important ones, though the threshold of being important is unknown. Therefore, we sample different thresholds $\tau$ in each iteration to give the model more flexibility to learn the importance. Then the above PMI can be written as
% \begin{equation}\label{eq:f_s}
%     f_S= \log \frac{P(\mathcal{R}| E)}{P(\mathcal{R})} 
%     \approx \int_{\tau\in[0,1]} P(\mathcal R_{\tau},S_{M_\tau},E)\cdot \log{\frac{P_{MLM}(\mathcal R_\tau|S_{M_\tau},E)}{P_{MLM}(\mathcal R_\tau|S_{M_\tau})}} \mathrm{d} \tau
% \end{equation}
\begin{equation}\label{eq:f_s}
\begin{aligned}
    f_S= & \log \frac{P(\mathcal{R}| E)}{P(\mathcal{R})} \\
    \approx & \int_{0}^{1} P(\tau)\cdot \log{\frac{P_{MLM}(\mathcal R_\tau|\mathcal G_{M_\tau},E)}{P_{MLM}(\mathcal R_\tau|\mathcal G_{M_\tau})}} \mathrm{d} \tau
\end{aligned}
\end{equation}
where $\tau$ denotes the ratio of tokens in $\mathcal{G}$ to be considered as important (e.g. $\tau=0.1$ means $\mathcal{R}_\tau$ includes the tokens with top 10\% high saliency scores in $\mathcal G$)
, and $P(\tau)$ is the distribution of sampling $\tau$, which can be implemented by any distribution that focuses on different thresholds (e.g. the uniform distribution from 0 to 0.3), $\mathcal R_{\tau}$ and $\mathcal G_{M_\tau}$ are the masked rationale $\mathcal{R}$ and the remaining text under the threshold $\tau$ (an illustration of constructing $G_{M_\tau}$ can be found in Appendix~\ref{append:serialize}).

\textbf{Faithfulness to predictions.} In addition to faithfulness to important inputs, we also encourage the generated explanations to be faithful to the outputs. Similarly, for faithfulness to predictions, we leverage the PMI between explanation $E$ and the predicted label $\hat y$ as a measurement as
\begin{equation}\label{eq:f_f}
    % f_F=\text{PMI}(E, \hat{y}) = \log \frac{P(\hat y, E)}{P(\hat y)P(E)} = \log \frac{P(\hat y| E)}{P(\hat y)} 
    f_F=\text{PMI}(E, \hat{y}) = \log \frac{P(\hat y| E)}{P(\hat y)} 
\end{equation}
where $\hat y$ denotes the textual form of the output prediction. The calculation of $P(\hat y| E)$ and $P(\hat y)$ is also implemented with a pre-trained language model.

\textbf{Brevity.} Since the above objective of faithfulness encourages $E$ to be informative, which may result in generating long and redundant explanations, which bring difficulty for humans understanding. Therefore, we also encourage the generated explanation $E$ to be concise, as
\begin{equation}\label{eq:f_b}
f_B=\frac{|E|}{|\mathcal G|}
\end{equation}
where $|E|$ and $|\mathcal G|$ denotes the length of $E$ and $\mathcal G$. 

Combining these measurements, we are essentially doing a multi-objective optimization problem, where we maximize $f_S$, $f_F$, and minimize $f_B$, which serves as our overall objective and be optimized with the following TAG expert iteration framework.

\subsubsection{TAG Explanation Expert Iteration} 
To effectively optimize the objectives to improve the \textit{Pseudo-Label Generator LLM}, we propose a TAG explanation iterative training method based on Expert Iteration \citep{dong2023raft, gulcehre2023reinforced}, as shown in Fig.~\ref{fig:method}~(b). Specifically, the training is composed of a closed loop of \textit{TAG explanation quality measuring}, \textit{high-quality TAG explanation selecting}, and \textit{TAG explanation updating}, as introduced in details as follows: 
\begin{enumerate}
    \item[(1)] TAG explanation quality measuring. Aligned with our training objectives, we calculate the scores $f_S$, $f_F$ and $f_B$ of the generated explanation pseudo-label $E$ based on Eq.~\ref{eq:f_s}, Eq.~\ref{eq:f_f} and Eq.~\ref{eq:f_b}.
    \item[(2)] High-quality TAG explanation selection.  Among all generated explanations, a subset of high-quality explanations is filtered from all candidates with customizable criteria to balance $f_S$, $f_F$, and $f_B$, such as weighted sum and top-k. 
    \item[(3)] TAG explanation updating. The selected high-quality explanations are used to fine-tune the \textit{Pseudo-Label Generator LLM}. Then the model generates a new batch of explanation candidates, and it goes back to step (1).
\end{enumerate}
Such three steps form a closed loop, allowing us to iteratively increase the performance of the model. Finally, we got the \textit{Pseudo-Label Generator LLM} fine-tuned to generate faithful and brief explanations with input as the saliency paragraph.

\subsection{End-to-End Explainer Training via Knowledge Distillation}\label{sec:distill}

The TAG Explanation Expert Iteration process gives us a pseudo-label generator model that experts in generating high-quality explanations based on saliency-based explanations. However, our goal is to have an end-to-end explainer model that can generate natural language explanations based on the raw input and its prediction. Therefore, after fine-tuning, we distill the whole pipeline to the \textit{Explainer LLM}. The distillation is conducted by accumulating a dataset of filtered high-quality candidates during the expert iteration process, and using this dataset to fine-tune the \textit{Explainer LLM}. The \textit{Explainer LLM} serves as the product of our framework, an end-to-end post-hoc explainer for the model $f$.

\section{Experiments}
\subsection{Experimental Setup}
\textbf{Datasets.} 
% We use three real-world TAG datasets, including two citation networks (Cora and DBLP) \textcolor{blue}{cite}, and one E-commerce co-purchasing network (Book-history \textcolor{blue}{cite}), to evaluate the performance of our method. Cora is a widely used citation network where each node represents a research paper, and edges represent citation links between the papers. It consists of 2,708 papers and 5,429 citation links, where each paper is assigned to one of seven topics. DBLP is a larger citation network with \textcolor{blue}{??} papers and \textcolor{blue}{??} co-authorship edges, categorized by \textcolor{blue}{??} research areas for a multi-class classification task. The Book-history dataset is an E-commerce co-purchasing network, where the nodes represent books, and an edge between two nodes indicates that the books were frequently purchased together. Book-history consists of \textcolor{blue}{??} books with \textcolor{blue}{??} co-purchasing links, and each book is classified into \textcolor{blue}{??} categories.
We use three real-world TAG datasets, including two citation networks (Cora~\citep{yang2016revisiting} and DBLP~\citep{tang2008arnetminer}), and one E-commerce co-purchasing network (Book-History~\citep{yan2023comprehensive}), to evaluate the performance of our method. More details of the datasets are given in Appendix~\ref{append:dataset}.

\begin{table*}[t]
\centering
\small
% \resizebox{\textwidth}{!}{%
\begin{tabular}{@{}clccccc@{}}
\toprule
\multirow{2}{*}{Dataset} & \multirow{2}{*}{Method} &
\multicolumn{5}{c}{Metrics} \\ \cmidrule(l){3-7} 
 &  & Simul. ($\uparrow$) & PMI-10\% ($\uparrow$) & PMI-20\% ($\uparrow$) & PMI-30\% ($\uparrow$) & Brevity ($\downarrow$) \\ \midrule

 \multirow{5}{*}{DBLP} 
 & LLaMA3.1 8B       & 0.63 & 0.139 & 0.109 & 0.077 & 0.394 \\
 & GPT-3.5 Turbo   & 0.71 & 0.136 & \textbf{0.110} & 0.084 & 0.403 \\
 & GPT-4o          & 0.82 & 0.142 & 0.101 & 0.085 & 0.385 \\
 & SMV             & 0.76 & 0.139 & 0.098 & 0.082 & 0.419 \\
 & $\text{GraphNarrator}_\text{LLaMA3.1 8B}$    & \textbf{0.95} & \textbf{0.155} & 0.108 & \textbf{0.085} & \textbf{0.354} \\ \midrule
 
\multirow{5}{*}{Cora} 
 & LLaMA3.1 8B       & 0.78 & 0.335 & 0.278 & 0.199 & 0.600 \\
 & GPT-3.5 Turbo   & 0.83 & 0.340 & 0.281 & 0.213 & 0.318 \\
 & GPT-4o          & 0.95 & 0.414 & 0.284 & 0.225 & 0.357 \\
 & SMV             & 0.88 & 0.359 & 0.267 & 0.217 & 0.431 \\
 & $\text{GraphNarrator}_\text{LLaMA3.1 8B}$    & \textbf{0.97} & \textbf{0.418} & \textbf{0.290} & \textbf{0.227} & \textbf{0.315} \\ \midrule

\multirow{5}{*}{Book-History} 
 & LLaMA3.1 8B       & 0.79 & 0.465 & \textbf{0.390} & 0.281 & 0.735 \\
 & GPT-3.5 Turbo   & 0.83 & 0.436 & 0.361 & 0.270 & 0.853 \\
 & GPT-4o          & 0.89 & 0.456 & 0.313 & 0.240 & 0.768 \\
 & SMV             & 0.87 & 0.441 & 0.320 & 0.257 & 0.836 \\
 & $\text{GraphNarrator}_\text{LLaMA3.1 8B}$    & \textbf{0.96} & \textbf{0.533} & 0.374 & \textbf{0.291} & \textbf{0.506} \\ 
\bottomrule
\end{tabular}
\caption{\textbf{Automatic evaluation} of natural language explanations quality generated by different methods. Best results are bolded.}
\vspace{-3mm}
\label{tab:main}
\end{table*}
\textbf{Compared Methods.} To our best knowledge, no existing method are designed to generate natural language explanations for graph learning. The most relevant method is SMV \citep{feldhus2022saliency}, which verbalizes saliency map explanations for text classification models. To evaluate the effectiveness of GraphNarrator, we compare it with various most advanced LLMs to generate explanations in a zero-shot manner given the input subgraph and the model prediction. We benchmarked our method with the most advanced LLMs, including GPT-4o, GPT-3.5 turbo, LLaMA 3.1, and SMV method based on GPT-4o (denoted as SMV in the results).

\textbf{Automatic Evaluation Metrics.} Automatic evaluation is applied to evaluate the faithfulness and brevity of explanations. Following previous research \citep{padmakumar2021unsupervised,  li2020evaluating}, \textit{Pointwise Mutual Information} (PMI) and \textit{Simulatability} (Simul.) are used as indicators for faithfulness. PMI \citep{padmakumar2021unsupervised, chen2022rev, colombo2022infolm, darrin2024cosmic} measures the mutual information between the generated explanations and the important regions in the input text. The top 10\%, 20\%, and 30\% important tokens are used as references for the calculation of PMI, denoted as PMI-10\%, PMI-20\%, and PMI-30\% in the result tables. \textit{Simulatability} \citep{sushil2018patient, sia2023logical, li2020evaluating, pruthi2022evaluating} measures the accuracy of the model prediction can be correctly inferred from the explanation. For \textit{Brevity}, the average ratio of explanation length and input length is used as an indicator. 

\textbf{Human Evaluation Setting.} We conduct a human annotation to investigate how human readers view the quality of explanations from different methods. For each dataset, we sample a split of 50 data points per method. We recruit three annotators with knowledge in computational linguistics (at least undergraduate level). Given the explanations, the annotators are asked to rate on a scale of 1-7 whether the explanations were (1) easy to understand, (2) insightful for the
underlying decision process, (3) informative in preserving graph semantics, (4) informative in preserving graph structures.

\textbf{Implementation Details.} 
Our TAG model to explain is implemented with the commonly used Bert+GNN pipeline. We use bert-base-uncased as the text encoder and 2-layer SAGE as the GNN backbone. For each dataset we train the TAG model until converged with a learning rate of 1e-3 and batch size of 500. We used the GPT-4o-mini-2024-07-18 model with our prompt for TAG explanation for generating explanation pseudo label candidates (the details of prompts are given in Appendix~\ref{sec:prompt}), applying one-shot learning for consistency. In the masked token prediction part, we utilized the gemma2-2b-it model to estimate the conditional probability distribution. We applied a balanced configuration for the three objectives (we select pseudo-label candidates whose three scores are all among the top 50\% of all candidates generated in that iteration). Finally, we used the fine-tuned LLaMA-3.1-8b as the base student model for knowledge distillation using the LoRA technique. More implementation details are given in Appendix~\ref{sec:impplement}.
\subsection{Explanation Quality}

\begin{table}[t]
\centering
\small
\begin{tabular}{@{}clcccc@{}}
\toprule
\multirow{2}{*}{Dataset} & \multirow{2}{*}{Method} & \multicolumn{4}{c}{Metrics} \\ \cmidrule(l){3-6} 
 &  & EU & DMI & SI & SeI \\ \midrule

\multirow{3}{*}{DBLP} 
 & GPT-4o         &  4.2 & 4.5 & 3.6 & 3.6  \\
 & SMV            & 3.8  &  4.1 &  3.4  &  3.7 \\
 & GraphNarrator    & \textbf{4.5}  & \textbf{4.8} &  \textbf{5.4}  &  \textbf{4.3} \\ \midrule
 
\multirow{3}{*}{Cora} 
 & GPT-4o         &  4.7 & 4.0  & 4.7  & 4.3  \\
 & SMV            & 4.2  & \textbf{5.2}  & 3.2  &  3.6 \\
 & GraphNarrator    &  \textbf{4.8} & 4.6  & \textbf{5.2}  & \textbf{5.0}  \\ \midrule

\multirow{3}{*}{History} 
 & GPT-4o         & 4.9 & 4.6 & 3.9 &  3.9  \\
 & SMV            & 4.8 & 4.7 & 3.9 &  4.1 \\
 & GraphNarrator   & \textbf{5.0} & \textbf{5.1} & \textbf{5.4} &  \textbf{5.3} \\ 

\bottomrule
\end{tabular}
\caption{\textbf{Human evaluation}. Best results are bolded. \textbf{EU}: Easy to Understand. \textbf{DMI}: Insightful in explaining the decision-making process. \textbf{SI}: Structural Informative. \textbf{SeI}: Semantic Informative.}
\vspace{-3mm}
\label{tab:human}
\end{table}

\textbf{Automatic Evaluation.} We use automatic methods to evaluate the faithfulness and brevity of generated explanations. Our experimental results, shown in Table~\ref{tab:main}, demonstrate that GraphNarrator consistently performs well in generating high-quality explanations. Specifically, GraphNarrator shows an 8.2\% average improvement over the second-best performer in the PMI-10\% metric over three datasets, highlighting its effectiveness at capturing the most important information for model decisions. In terms of simulatability, GraphNarrator outperforms all baseline methods by 8.6\%, achieving a simulatability score of 0.95 across all three datasets, significantly higher than other methods, proving highly faithful to model predictions. For the brevity metric, GraphNarrator is 13.4\% better than the second-best performer, effectively balancing conciseness and accuracy. Across all three datasets, GraphNarrator generates relatively compact explanations while maintaining high simulatability scores. These results demonstrate that GraphNarrator successfully navigates the inherent trade-offs among PMI, simulatability, and brevity. It consistently produces explanations that align closely with model predictions while remaining concise, further enhancing their interpretability. This balance highlights GraphNarrator’s strength in delivering both faithful and interpretable explanations. The improvement on Cora dataset is relatively low due to the low data quality brought OCR-based collection.

\begin{table*}[ht]
\vspace{-5mm}
\centering
\small
% \resizebox{\textwidth}{!}{%
\begin{tabular}{@{}lccccc@{}}
\toprule
\multirow{2}{*}{Method} &
\multicolumn{5}{c}{Metrics} \\ \cmidrule(l){2-6} 
 & Simul. ($\uparrow$) & PMI-10\% ($\uparrow$) & PMI-20\% ($\uparrow$) & PMI-30\% ($\uparrow$) & Brevity ($\downarrow$) \\ \midrule

GraphNarrator    & 0.97 & 0.418 & 0.290 & 0.227 & 0.315 \\ \midrule
w/o $f_S$       & 0.98 & \cellcolor{red!20}0.407 & \cellcolor{red!20}0.298 & \cellcolor{red!20}0.213 & 0.304 \\
w/o $f_F$       & \cellcolor{red!20}0.90 & 0.419 & 0.311 & 0.241 & 0.315 \\
w/o $f_B$       & 0.96 & 0.432 & 0.327 & 0.239 & \cellcolor{red!20}0.361 \\
w/o saliency    & 0.97 & \cellcolor{red!20}0.402 & \cellcolor{red!20}0.284 & \cellcolor{red!20}0.218 & 0.343 \\ 
% w/o Expert Iteration       & \cellcolor{red!20}0.414 & \cellcolor{red!20}0.284 & \cellcolor{red!20}0.225 & \cellcolor{red!20}0.95 & \cellcolor{red!20}0.357 \\ 
\bottomrule
\end{tabular}%
% }
\caption{Results of ablation study. w/o $f_S$, $f_F$, $F_B$ denotes removing the corresponding objective for training. Cells highlighted represent metrics where the performance is \textbf{expected to drop} when the corresponding component is removed.}
\label{tab:2}
\vspace{-3mm}
\end{table*}

\textbf{Human Evaluation.}
Results of human evaluation are given in Table~\ref{tab:human}. Results reveal our approach demonstrates strong performance across multiple dimensions of human evaluation. In particular, it achieves the best overall results in terms of ease of understanding, insightfulness in the explanation process, and the preservation of both semantic and structural information. When compared with GPT-4o, our method yields notable improvements—approximately 33.7\% in structural informativeness and 23.9\% in semantic informativeness. Moreover, its ease of understanding is on par with both GPT-4o and SMV, a benefit likely derived from the robust large language model backbone underlying our approach. 
\subsection{Pseudo-Label Self-Improvement}
\begin{figure}[t]
\vspace{-3mm}
    \centering
    \begin{subfigure}[b]{0.50\textwidth}
        \centering
        \includegraphics[width=\textwidth]{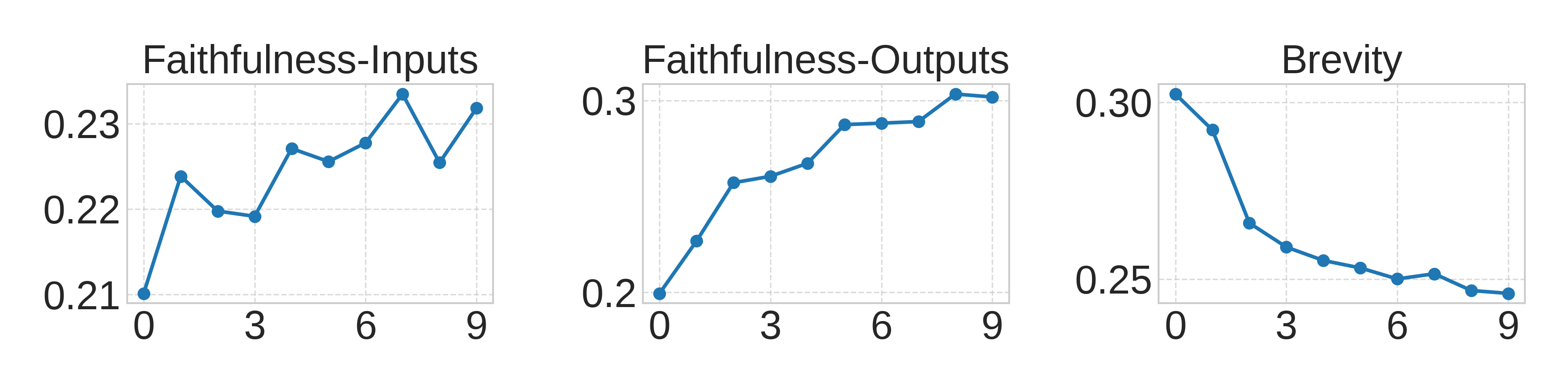}
        \vspace{-8mm} 
        \caption{Cora dataset.}
        \label{fig:sub1}
    \end{subfigure}
    \begin{subfigure}[b]{0.50\textwidth}
        \centering
        \includegraphics[width=\textwidth]{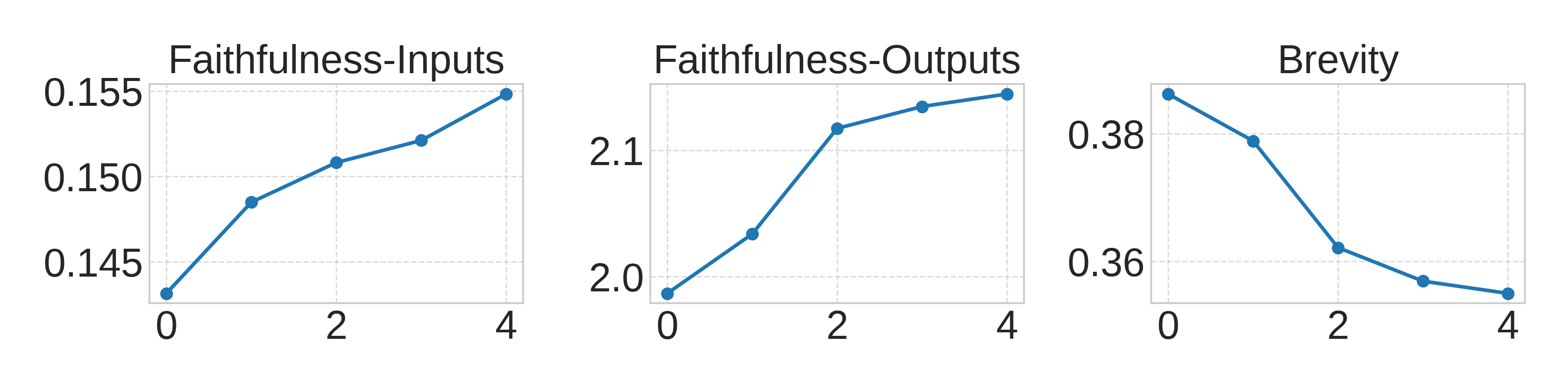}
        \vspace{-8mm} 
        \caption{DBLP dataset.}
        \label{fig:sub2}
    \end{subfigure}
    \vspace{-3mm} 
    \begin{subfigure}[b]{0.50\textwidth}
        \centering
        \includegraphics[width=\textwidth]{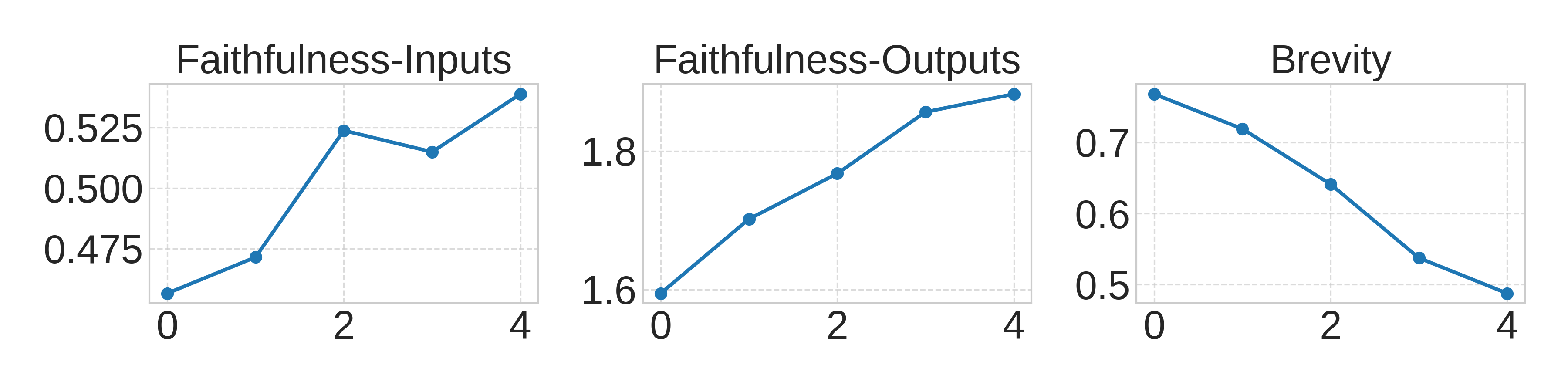}
        \vspace{-8mm} 
        \caption{Book-History dataset.}
        \label{fig:sub3}
    \end{subfigure}
    \vspace{-3mm}
    \caption{The change of three pseudo-label quality scores in the TAG explanation expert iteration process, w.r.t number of training iteration.}
    \vspace{-3mm}
    \label{fig:curve}
\end{figure}
 
\textbf{Effectiveness of TAG Explanation Expert Iteration.} The training score curve in our iterative training (Expert Iteration) of the pseudo-label generator LLM is illustrated in Figure \ref{fig:curve}. With an increasing number of iterations, both faithfulness to important inputs and faithfulness to predictions exhibit a overall upward trend, while brevity shows a gradual decline. 
This iterative learning process underscores our TAG explanation expert iteration's effectiveness in improving the explanation quality.
It is worth highlighting that during each iteration of the expert iteration process, we select only 50 high-quality samples. Despite this small sample number, our TAG explanation expert iteration consistently enhances faithfulness to both important inputs and predictions. This demonstrates that the process is highly efficient, as steady improvements in performance are achieved by using just a small and high-quality set of high-quality samples in each iteration. We also evaluated the quality of pseudo-labels to verify the effectiveness of the Expert Iteration pipeline, as given in Appendix~\ref{append:pseudo-quality}. A study of different pseudo-label selection strategies is given in  Appendix~\ref{appendix:strategy}. The performance improvement of the explainer LLM after fine-tuning with generated high-quality pseudo-labels are given in Appendix~\ref{append:Distill}.

\subsection{Ablation Study}

In the ablation study, we removed $f_S$, $f_F$, $f_B$ and Expert Iteration to test their corresponding effectiveness, as shown in Table~\ref{tab:2}. The results highlight the effectiveness of different components of the proposed framework. Removing the saliency optimization objective (w/o $f_S$) decreases PMI scores, proving its importance for faithfulness. Removing the fidelity objective (w/o $f_F$) results in lower simulatability, and removing the brevity optimization objective (w/o $f_B$). Notably, sometimes when one objective is removed, the other two often improve. This is because the three objectives inherently involve trade-offs—removing one allows the remaining two to be optimized within a larger space, free from the constraints imposed by the removed objective, naturally leading to better performance in those areas.

\section{Conclusion}

In this paper, we present GraphNarrator, a model-agnostic post-hoc explainer to generate natural language explanations for TAG learning models. GraphNarrator fine-tunes a generative language model as an explanation generator with pseudo-labels derived from saliency-based explanations. Through iterative self-training, we improve the quality of generated explanation pseudo-labels, ensuring the explanation generator can be trained with high-quality data. Our extensive experiments demonstrate the effectiveness of GraphNarrator.

\section{Acknowledgments}
 This work was supported by the National Science Foundation (NSF) Grant No. 2414115, No. 2403312, No. 2007716, No. 2007976, No. 1942594, No. 1907805, NIH R01AG089806, and NIH R01CA297856.

\section*{Limitation}
The backbone of GraphNarrator is based on LLMs, which may be more costly to do inference than saliency-based explanation methods. For long documents although KV-cache can bring significant improvements for inference, we still find for a few extremely large subgraphs, the inference time could exceed 2 mins.

\bibliography{acl}

\begin{thebibliography}{49}
\providecommand{\natexlab}[1]{#1}

\bibitem[{Agarwal et~al.(2024)Agarwal, Tanneru, and Lakkaraju}]{agarwal2024faithfulness}
Chirag Agarwal, Sree~Harsha Tanneru, and Himabindu Lakkaraju. 2024.
\newblock Faithfulness vs. plausibility: On the (un) reliability of explanations from large language models.
\newblock \emph{arXiv preprint arXiv:2402.04614}.

\bibitem[{Bach et~al.(2015)Bach, Binder, Montavon, Klauschen, M{\"u}ller, and Samek}]{bach2015pixel}
Sebastian Bach, Alexander Binder, Gr{\'e}goire Montavon, Frederick Klauschen, Klaus-Robert M{\"u}ller, and Wojciech Samek. 2015.
\newblock On pixel-wise explanations for non-linear classifier decisions by layer-wise relevance propagation.
\newblock \emph{PloS one}, 10(7):e0130140.

\bibitem[{Baldassarre and Azizpour(2019)}]{baldassarre2019explainability}
Federico Baldassarre and Hossein Azizpour. 2019.
\newblock Explainability techniques for graph convolutional networks.
\newblock In \emph{International Conference on Machine Learning (ICML) Workshops, 2019 Workshop on Learning and Reasoning with Graph-Structured Representations}.

\bibitem[{Cambria et~al.(2023)Cambria, Malandri, Mercorio, Mezzanzanica, and Nobani}]{cambria2023survey}
Erik Cambria, Lorenzo Malandri, Fabio Mercorio, Mario Mezzanzanica, and Navid Nobani. 2023.
\newblock A survey on xai and natural language explanations.
\newblock \emph{Information Processing \& Management}, 60(1):103111.

\bibitem[{Camburu et~al.(2018)Camburu, Rockt{\"a}schel, Lukasiewicz, and Blunsom}]{camburu2018snli}
Oana-Maria Camburu, Tim Rockt{\"a}schel, Thomas Lukasiewicz, and Phil Blunsom. 2018.
\newblock e-snli: Natural language inference with natural language explanations.
\newblock \emph{Advances in Neural Information Processing Systems}, 31.

\bibitem[{Chen et~al.(2022)Chen, Brahman, Ren, Ji, Choi, and Swayamdipta}]{chen2022rev}
Hanjie Chen, Faeze Brahman, Xiang Ren, Yangfeng Ji, Yejin Choi, and Swabha Swayamdipta. 2022.
\newblock Rev: information-theoretic evaluation of free-text rationales.
\newblock \emph{arXiv preprint arXiv:2210.04982}.

\bibitem[{Colombo et~al.(2022)Colombo, Clavel, and Piantanida}]{colombo2022infolm}
Pierre Jean~A Colombo, Chlo{\'e} Clavel, and Pablo Piantanida. 2022.
\newblock Infolm: A new metric to evaluate summarization \& data2text generation.
\newblock In \emph{Proceedings of the AAAI conference on artificial intelligence}, volume~36, pages 10554--10562.

\bibitem[{Darrin et~al.(2024)Darrin, Formont, Cheung, and Piantanida}]{darrin2024cosmic}
Maxime Darrin, Philippe Formont, Jackie Chi~Kit Cheung, and Pablo Piantanida. 2024.
\newblock Cosmic: Mutual information for task-agnostic summarization evaluation.
\newblock In \emph{Proceedings of the 62nd Annual Meeting of the Association for Computational Linguistics (Volume 1: Long Papers)}, pages 12696--12717.

\bibitem[{Dong et~al.(2023)Dong, Xiong, Goyal, Zhang, Chow, Pan, Diao, Zhang, Shum, and Zhang}]{dong2023raft}
Hanze Dong, Wei Xiong, Deepanshu Goyal, Yihan Zhang, Winnie Chow, Rui Pan, Shizhe Diao, Jipeng Zhang, Kashun Shum, and Tong Zhang. 2023.
\newblock Raft: Reward ranked finetuning for generative foundation model alignment.
\newblock \emph{arXiv preprint arXiv:2304.06767}.

\bibitem[{Feldhus et~al.(2022)Feldhus, Hennig, Nasert, Ebert, Schwarzenberg, and M{\"o}ller}]{feldhus2022saliency}
Nils Feldhus, Leonhard Hennig, Maximilian~Dustin Nasert, Christopher Ebert, Robert Schwarzenberg, and Sebastian M{\"o}ller. 2022.
\newblock Saliency map verbalization: Comparing feature importance representations from model-free and instruction-based methods.
\newblock \emph{arXiv preprint arXiv:2210.07222}.

\bibitem[{Gat et~al.(2023)Gat, Calderon, Feder, Chapanin, Sharma, and Reichart}]{gat2023faithful}
Yair Gat, Nitay Calderon, Amir Feder, Alexander Chapanin, Amit Sharma, and Roi Reichart. 2023.
\newblock Faithful explanations of black-box nlp models using llm-generated counterfactuals.
\newblock \emph{arXiv preprint arXiv:2310.00603}.

\bibitem[{Gulcehre et~al.(2023)Gulcehre, Paine, Srinivasan, Konyushkova, Weerts, Sharma, Siddhant, Ahern, Wang, Gu et~al.}]{gulcehre2023reinforced}
Caglar Gulcehre, Tom~Le Paine, Srivatsan Srinivasan, Ksenia Konyushkova, Lotte Weerts, Abhishek Sharma, Aditya Siddhant, Alex Ahern, Miaosen Wang, Chenjie Gu, et~al. 2023.
\newblock Reinforced self-training (rest) for language modeling.
\newblock \emph{arXiv preprint arXiv:2308.08998}.

\bibitem[{Hamilton et~al.(2017)Hamilton, Ying, and Leskovec}]{hamilton2017inductive}
Will Hamilton, Zhitao Ying, and Jure Leskovec. 2017.
\newblock Inductive representation learning on large graphs.
\newblock \emph{Advances in neural information processing systems}, 30.

\bibitem[{He et~al.(2023)He, Bresson, Laurent, Perold, LeCun, and Hooi}]{he2023harnessing}
Xiaoxin He, Xavier Bresson, Thomas Laurent, Adam Perold, Yann LeCun, and Bryan Hooi. 2023.
\newblock Harnessing explanations: Llm-to-lm interpreter for enhanced text-attributed graph representation learning.
\newblock \emph{arXiv preprint arXiv:2305.19523}.

\bibitem[{Hu et~al.(2024)Hu, Lei, Zhang, Pan, Ling, and Zhao}]{hu2024grag}
Yuntong Hu, Zhihan Lei, Zheng Zhang, Bo~Pan, Chen Ling, and Liang Zhao. 2024.
\newblock Grag: Graph retrieval-augmented generation.
\newblock \emph{arXiv preprint arXiv:2405.16506}.

\bibitem[{Huang et~al.(2022)Huang, Yamada, Tian, Singh, and Chang}]{huang2022graphlime}
Qiang Huang, Makoto Yamada, Yuan Tian, Dinesh Singh, and Yi~Chang. 2022.
\newblock Graphlime: Local interpretable model explanations for graph neural networks.
\newblock \emph{IEEE Transactions on Knowledge and Data Engineering}, 35(7):6968--6972.

\bibitem[{Kipf and Welling(2016)}]{kipf2016semi}
Thomas~N Kipf and Max Welling. 2016.
\newblock Semi-supervised classification with graph convolutional networks.
\newblock \emph{arXiv preprint arXiv:1609.02907}.

\bibitem[{Kojima et~al.(2022)Kojima, Gu, Reid, Matsuo, and Iwasawa}]{kojima2022large}
Takeshi Kojima, Shixiang~Shane Gu, Machel Reid, Yutaka Matsuo, and Yusuke Iwasawa. 2022.
\newblock Large language models are zero-shot reasoners.
\newblock \emph{Advances in neural information processing systems}, 35:22199--22213.

\bibitem[{Kroeger et~al.(2023)Kroeger, Ley, Krishna, Agarwal, and Lakkaraju}]{kroeger2023large}
Nicholas Kroeger, Dan Ley, Satyapriya Krishna, Chirag Agarwal, and Himabindu Lakkaraju. 2023.
\newblock Are large language models post hoc explainers?
\newblock \emph{arXiv preprint arXiv:2310.05797}.

\bibitem[{Lei et~al.(2016)Lei, Barzilay, and Jaakkola}]{lei2016rationalizing}
Tao Lei, Regina Barzilay, and Tommi Jaakkola. 2016.
\newblock Rationalizing neural predictions.
\newblock \emph{arXiv preprint arXiv:1606.04155}.

\bibitem[{Li et~al.(2020)Li, Liu, Li, Li, Huang, and Shi}]{li2020evaluating}
Jierui Li, Lemao Liu, Huayang Li, Guanlin Li, Guoping Huang, and Shuming Shi. 2020.
\newblock Evaluating explanation methods for neural machine translation.
\newblock \emph{arXiv preprint arXiv:2005.01672}.

\bibitem[{Liu et~al.(2024)Liu, Wang, Wang, Deng, Zhang, Wang, and Li}]{liu2024enhancing}
Wei Liu, Haozhao Wang, Jun Wang, Zhiying Deng, Yuankai Zhang, Cheng Wang, and Ruixuan Li. 2024.
\newblock Enhancing the rationale-input alignment for self-explaining rationalization.
\newblock In \emph{2024 IEEE 40th International Conference on Data Engineering (ICDE)}, pages 2218--2230. IEEE.

\bibitem[{Liu et~al.(2023)Liu, Wang, Wang, Li, Deng, Zhang, and Qiu}]{liu2023d}
Wei Liu, Jun Wang, Haozhao Wang, Ruixuan Li, Zhiying Deng, Yuankai Zhang, and Yang Qiu. 2023.
\newblock D-separation for causal self-explanation.
\newblock \emph{Advances in Neural Information Processing Systems}, 36:43620--43633.

\bibitem[{Luo et~al.(2020)Luo, Cheng, Xu, Yu, Zong, Chen, and Zhang}]{luo2020parameterized}
Dongsheng Luo, Wei Cheng, Dongkuan Xu, Wenchao Yu, Bo~Zong, Haifeng Chen, and Xiang Zhang. 2020.
\newblock Parameterized explainer for graph neural network.
\newblock \emph{Advances in neural information processing systems}, 33:19620--19631.

\bibitem[{Martens et~al.(2023)Martens, Hinns, Dams, Vergouwen, and Evgeniou}]{martens2023tell}
David Martens, James Hinns, Camille Dams, Mark Vergouwen, and Theodoros Evgeniou. 2023.
\newblock Tell me a story! narrative-driven xai with large language models.
\newblock \emph{arXiv preprint arXiv:2309.17057}.

\bibitem[{Narang et~al.(2020)Narang, Raffel, Lee, Roberts, Fiedel, and Malkan}]{narang2020wt5}
Sharan Narang, Colin Raffel, Katherine Lee, Adam Roberts, Noah Fiedel, and Karishma Malkan. 2020.
\newblock Wt5?! training text-to-text models to explain their predictions.
\newblock \emph{arXiv preprint arXiv:2004.14546}.

\bibitem[{Ni et~al.(2019)Ni, Li, and McAuley}]{ni2019justifying}
Jianmo Ni, Jiacheng Li, and Julian McAuley. 2019.
\newblock Justifying recommendations using distantly-labeled reviews and fine-grained aspects.
\newblock In \emph{Proceedings of the 2019 conference on empirical methods in natural language processing and the 9th international joint conference on natural language processing (EMNLP-IJCNLP)}, pages 188--197.

\bibitem[{Padmakumar and He(2021)}]{padmakumar2021unsupervised}
Vishakh Padmakumar and He~He. 2021.
\newblock Unsupervised extractive summarization using pointwise mutual information.
\newblock \emph{arXiv preprint arXiv:2102.06272}.

\bibitem[{Pan et~al.(2024)Pan, Zhang, Zhang, Hu, and Zhao}]{pan2024distilling}
Bo~Pan, Zheng Zhang, Yifei Zhang, Yuntong Hu, and Liang Zhao. 2024.
\newblock Distilling large language models for text-attributed graph learning.
\newblock \emph{arXiv preprint arXiv:2402.12022}.

\bibitem[{Parcalabescu and Frank(2024)}]{parcalabescu2024measuring}
Letitia Parcalabescu and Anette Frank. 2024.
\newblock On measuring faithfulness or self-consistency of natural language explanations.
\newblock In \emph{Proceedings of the 62nd Annual Meeting of the Association for Computational Linguistics (Volume 1: Long Papers)}, pages 6048--6089.

\bibitem[{Pruthi et~al.(2022)Pruthi, Bansal, Dhingra, Soares, Collins, Lipton, Neubig, and Cohen}]{pruthi2022evaluating}
Danish Pruthi, Rachit Bansal, Bhuwan Dhingra, Livio~Baldini Soares, Michael Collins, Zachary~C Lipton, Graham Neubig, and William~W Cohen. 2022.
\newblock Evaluating explanations: How much do explanations from the teacher aid students?
\newblock \emph{Transactions of the Association for Computational Linguistics}, 10:359--375.

\bibitem[{Rajani et~al.(2019)Rajani, McCann, Xiong, and Socher}]{rajani2019explain}
Nazneen~Fatema Rajani, Bryan McCann, Caiming Xiong, and Richard Socher. 2019.
\newblock Explain yourself! leveraging language models for commonsense reasoning.
\newblock \emph{arXiv preprint arXiv:1906.02361}.

\bibitem[{Schwarzenberg et~al.(2019)Schwarzenberg, H{\"u}bner, Harbecke, Alt, and Hennig}]{schwarzenberg2019layerwise}
Robert Schwarzenberg, Marc H{\"u}bner, David Harbecke, Christoph Alt, and Leonhard Hennig. 2019.
\newblock Layerwise relevance visualization in convolutional text graph classifiers.
\newblock \emph{arXiv preprint arXiv:1909.10911}.

\bibitem[{Sia et~al.(2023)Sia, Belyy, Almahairi, Khabsa, Zettlemoyer, and Mathias}]{sia2023logical}
Suzanna Sia, Anton Belyy, Amjad Almahairi, Madian Khabsa, Luke Zettlemoyer, and Lambert Mathias. 2023.
\newblock Logical satisfiability of counterfactuals for faithful explanations in nli.
\newblock In \emph{Proceedings of the AAAI Conference on Artificial Intelligence}, volume~37, pages 9837--9845.

\bibitem[{{\v{S}}trumbelj and Kononenko(2014)}]{vstrumbelj2014explaining}
Erik {\v{S}}trumbelj and Igor Kononenko. 2014.
\newblock Explaining prediction models and individual predictions with feature contributions.
\newblock \emph{Knowledge and information systems}, 41:647--665.

\bibitem[{Sushil et~al.(2018)Sushil, {\v{S}}uster, Luyckx, and Daelemans}]{sushil2018patient}
Madhumita Sushil, Simon {\v{S}}uster, Kim Luyckx, and Walter Daelemans. 2018.
\newblock Patient representation learning and interpretable evaluation using clinical notes.
\newblock \emph{Journal of biomedical informatics}, 84:103--113.

\bibitem[{Tang et~al.(2008)Tang, Zhang, Yao, Li, Zhang, and Su}]{tang2008arnetminer}
Jie Tang, Jing Zhang, Limin Yao, Juanzi Li, Li~Zhang, and Zhong Su. 2008.
\newblock Arnetminer: extraction and mining of academic social networks.
\newblock In \emph{Proceedings of the 14th ACM SIGKDD international conference on Knowledge discovery and data mining}, pages 990--998.

\bibitem[{Vu and Thai(2020)}]{vu2020pgm}
Minh Vu and My~T Thai. 2020.
\newblock Pgm-explainer: Probabilistic graphical model explanations for graph neural networks.
\newblock \emph{Advances in neural information processing systems}, 33:12225--12235.

\bibitem[{Wei et~al.(2022)Wei, Wang, Schuurmans, Bosma, Xia, Chi, Le, Zhou et~al.}]{wei2022chain}
Jason Wei, Xuezhi Wang, Dale Schuurmans, Maarten Bosma, Fei Xia, Ed~Chi, Quoc~V Le, Denny Zhou, et~al. 2022.
\newblock Chain-of-thought prompting elicits reasoning in large language models.
\newblock \emph{Advances in Neural Information Processing Systems}, 35:24824--24837.

\bibitem[{Wiegreffe et~al.(2020)Wiegreffe, Marasovi{\'c}, and Smith}]{wiegreffe2020measuring}
Sarah Wiegreffe, Ana Marasovi{\'c}, and Noah~A Smith. 2020.
\newblock Measuring association between labels and free-text rationales.
\newblock \emph{arXiv preprint arXiv:2010.12762}.

\bibitem[{Yan et~al.(2023)Yan, Li, Long, Yan, Zhao, Zhuang, Yin, Zhang, Han, Sun et~al.}]{yan2023comprehensive}
Hao Yan, Chaozhuo Li, Ruosong Long, Chao Yan, Jianan Zhao, Wenwen Zhuang, Jun Yin, Peiyan Zhang, Weihao Han, Hao Sun, et~al. 2023.
\newblock A comprehensive study on text-attributed graphs: Benchmarking and rethinking.
\newblock \emph{Advances in Neural Information Processing Systems}, 36:17238--17264.

\bibitem[{Yang et~al.(2021)Yang, Liu, Xiao, Li, Lian, Agrawal, Singh, Sun, and Xie}]{yang2021graphformers}
Junhan Yang, Zheng Liu, Shitao Xiao, Chaozhuo Li, Defu Lian, Sanjay Agrawal, Amit Singh, Guangzhong Sun, and Xing Xie. 2021.
\newblock Graphformers: Gnn-nested transformers for representation learning on textual graph.
\newblock \emph{Advances in Neural Information Processing Systems}, 34:28798--28810.

\bibitem[{Yang et~al.(2016)Yang, Cohen, and Salakhudinov}]{yang2016revisiting}
Zhilin Yang, William Cohen, and Ruslan Salakhudinov. 2016.
\newblock Revisiting semi-supervised learning with graph embeddings.
\newblock In \emph{International conference on machine learning}, pages 40--48. PMLR.

\bibitem[{Ying et~al.(2019)Ying, Bourgeois, You, Zitnik, and Leskovec}]{ying2019gnnexplainer}
Zhitao Ying, Dylan Bourgeois, Jiaxuan You, Marinka Zitnik, and Jure Leskovec. 2019.
\newblock Gnnexplainer: Generating explanations for graph neural networks.
\newblock \emph{Advances in neural information processing systems}, 32.

\bibitem[{Yu et~al.(2019)Yu, Chang, Zhang, and Jaakkola}]{yu2019rethinking}
Mo~Yu, Shiyu Chang, Yang Zhang, and Tommi~S Jaakkola. 2019.
\newblock Rethinking cooperative rationalization: Introspective extraction and complement control.
\newblock \emph{arXiv preprint arXiv:1910.13294}.

\bibitem[{Yuan et~al.(2020)Yuan, Tang, Hu, and Ji}]{yuan2020xgnn}
Hao Yuan, Jiliang Tang, Xia Hu, and Shuiwang Ji. 2020.
\newblock Xgnn: Towards model-level explanations of graph neural networks.
\newblock In \emph{Proceedings of the 26th ACM SIGKDD international conference on knowledge discovery \& data mining}, pages 430--438.

\bibitem[{Zhang et~al.(2024)Zhang, Pan, Ling, Hu, and Zhao}]{zhang2024elad}
Yifei Zhang, Bo~Pan, Chen Ling, Yuntong Hu, and Liang Zhao. 2024.
\newblock Elad: Explanation-guided large language models active distillation.
\newblock \emph{arXiv preprint arXiv:2402.13098}.

\bibitem[{Zhang et~al.(2022)Zhang, Liu, Wang, Lu, and Lee}]{zhang2022protgnn}
Zaixi Zhang, Qi~Liu, Hao Wang, Chengqiang Lu, and Cheekong Lee. 2022.
\newblock Protgnn: Towards self-explaining graph neural networks.
\newblock In \emph{Proceedings of the AAAI Conference on Artificial Intelligence}, volume~36, pages 9127--9135.

\bibitem[{Zhou et~al.(2019)Zhou, Han, Yang, Liu, Wang, Li, and Sun}]{zhou2019gear}
Jie Zhou, Xu~Han, Cheng Yang, Zhiyuan Liu, Lifeng Wang, Changcheng Li, and Maosong Sun. 2019.
\newblock Gear: Graph-based evidence aggregating and reasoning for fact verification.
\newblock \emph{arXiv preprint arXiv:1908.01843}.

\end{thebibliography}

\appendix
\newpage

\newpage

\section{Qualitative Evaluation}
\begin{figure*}[hbt!]
    \centering
    \begin{subfigure}[b]{0.5\textwidth}
        \centering
        \includegraphics[width=\textwidth]{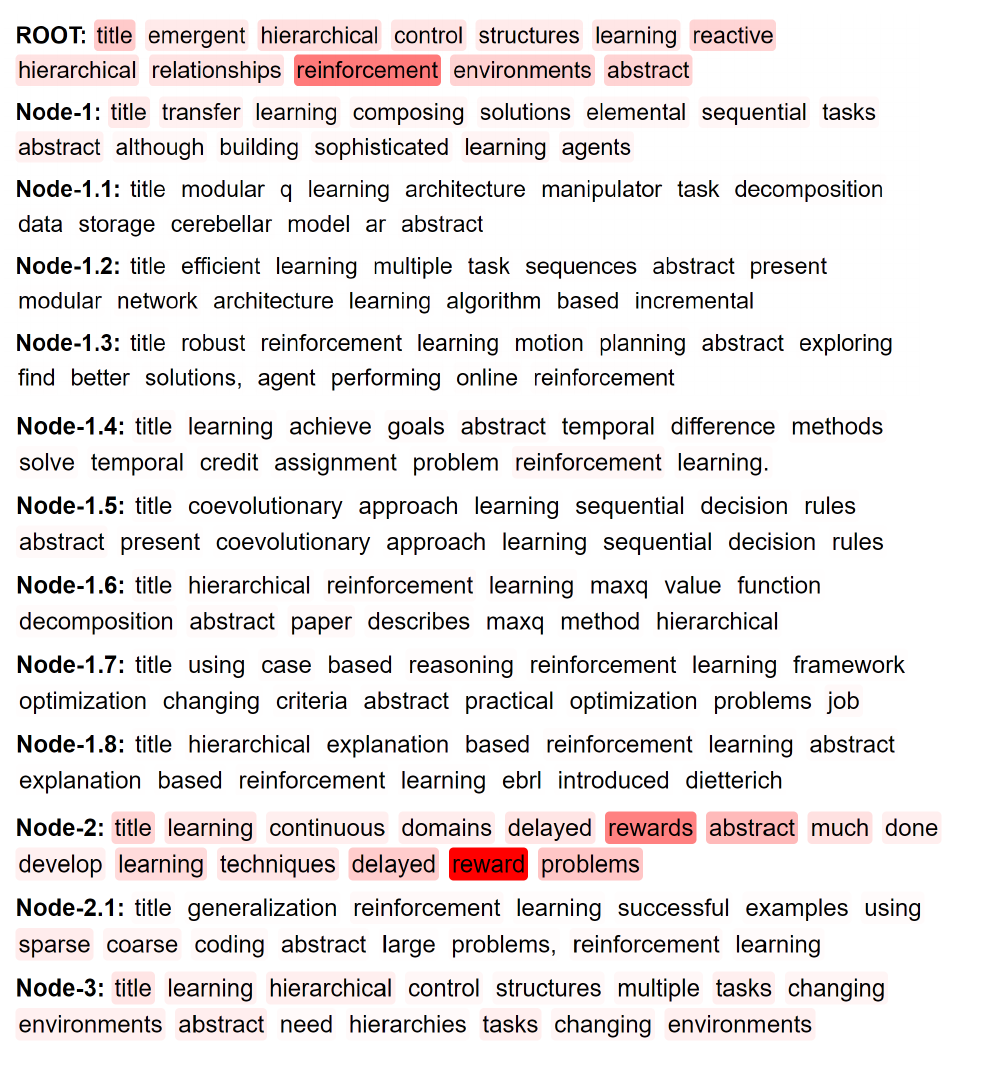}
        \vspace{-6mm} 
        \caption{Saliency-Based Explanation}
        \label{fig:visual-saliency-map}
    \end{subfigure}
    \begin{subfigure}[b]{0.48\textwidth}
        \centering
        \includegraphics[width=\textwidth]{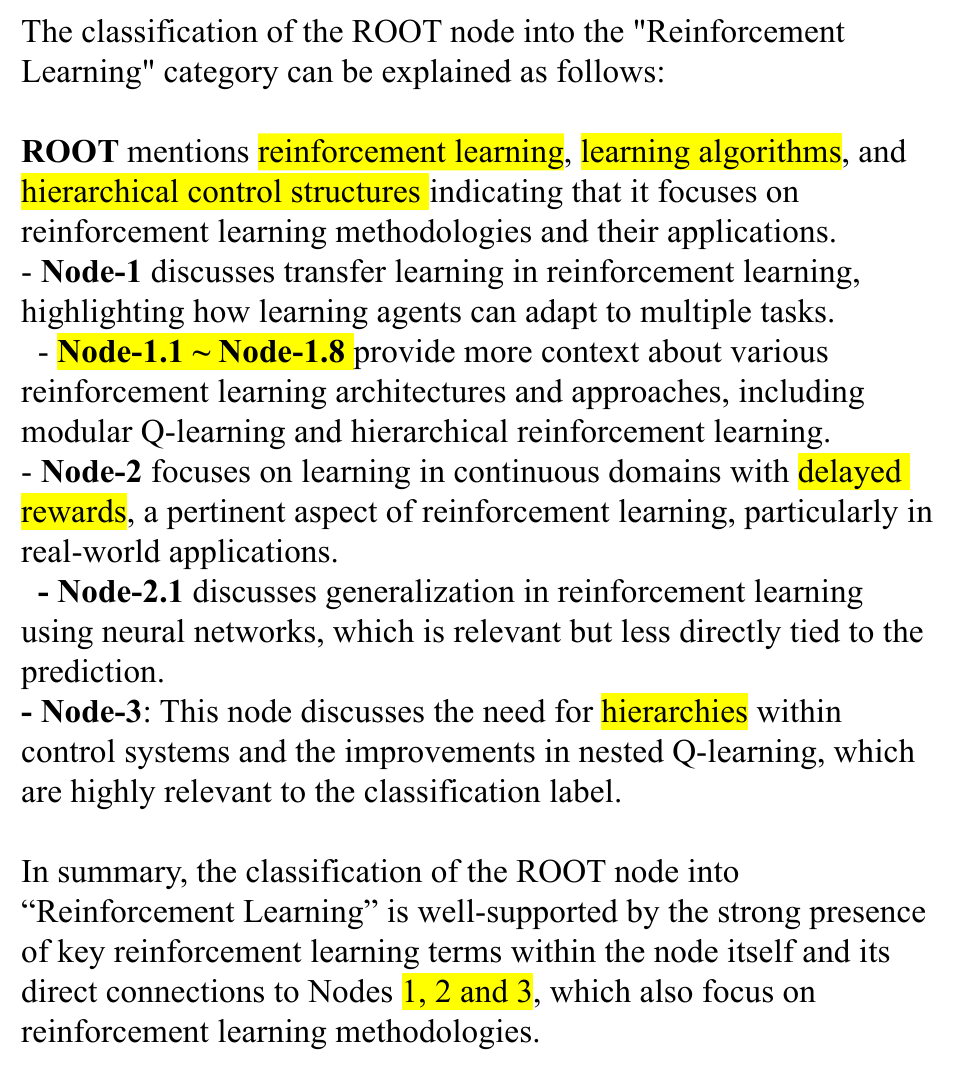}
        \vspace{-6mm} 
        \caption{GraphNarrator}
        \label{fig:visual-explanation}
    \end{subfigure}
    \caption{Visualization of a saliency-based explanation and a corresponding natural language explanation generated by GraphNarrator. In (a), red words indicate important terms, with darker red showing higher importance. We only visualized first 15 words in each paper due to the space limitation. In (b), yellow highlights reference high-saliency areas and emphasize that the explanation summarized key information.}
    \label{fig:visualization}
    \vspace{-2mm}
\end{figure*}

\begin{figure*}[hbt!]
\begin{center}
\includegraphics[width=1.0\textwidth]{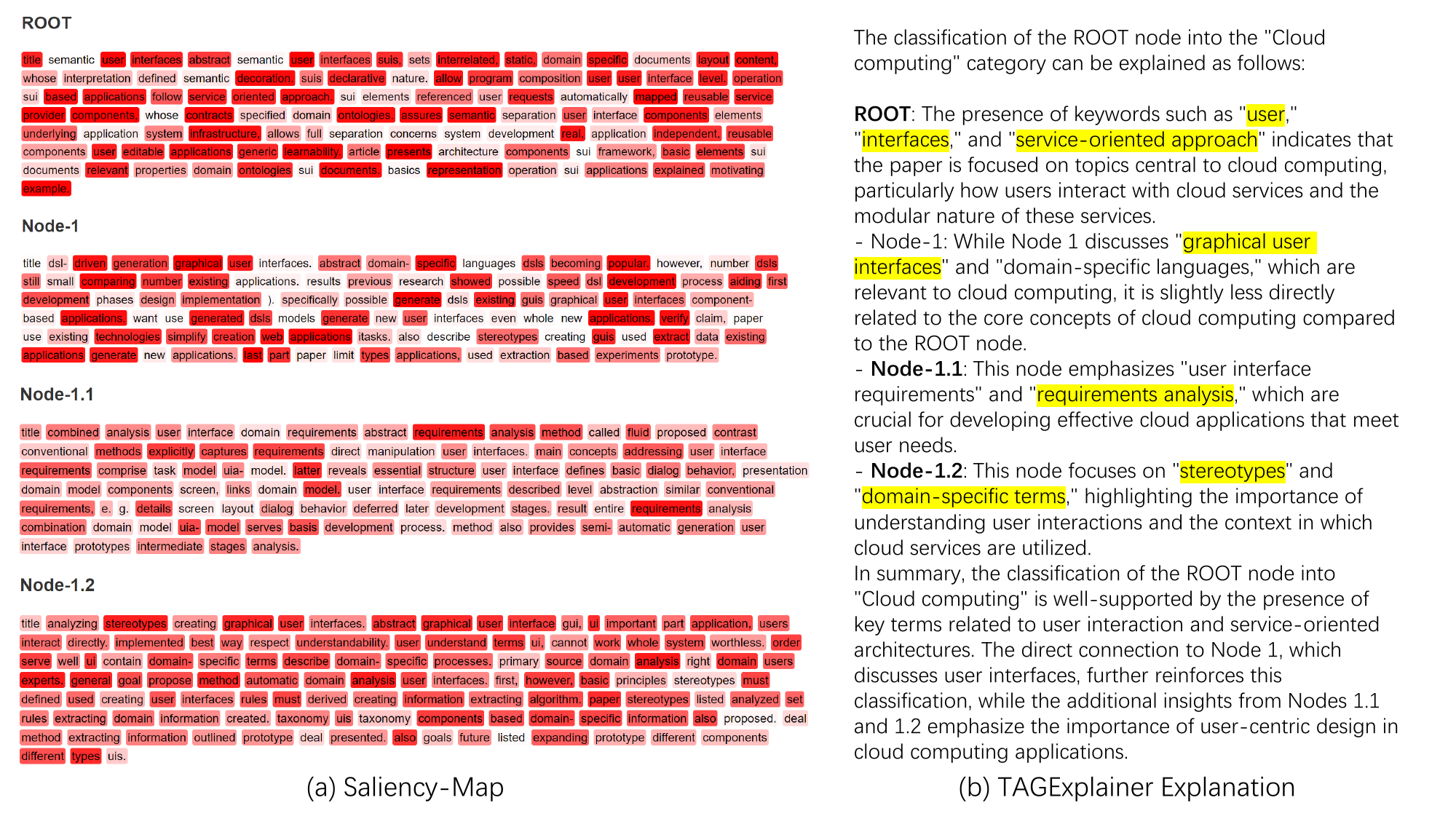}
\end{center}
\vspace{-3mm}
\caption{Example from DBLP}\label{fig:6}
\end{figure*}

\begin{figure*}[hbt!]
\begin{center}
\includegraphics[width=1.0\textwidth]{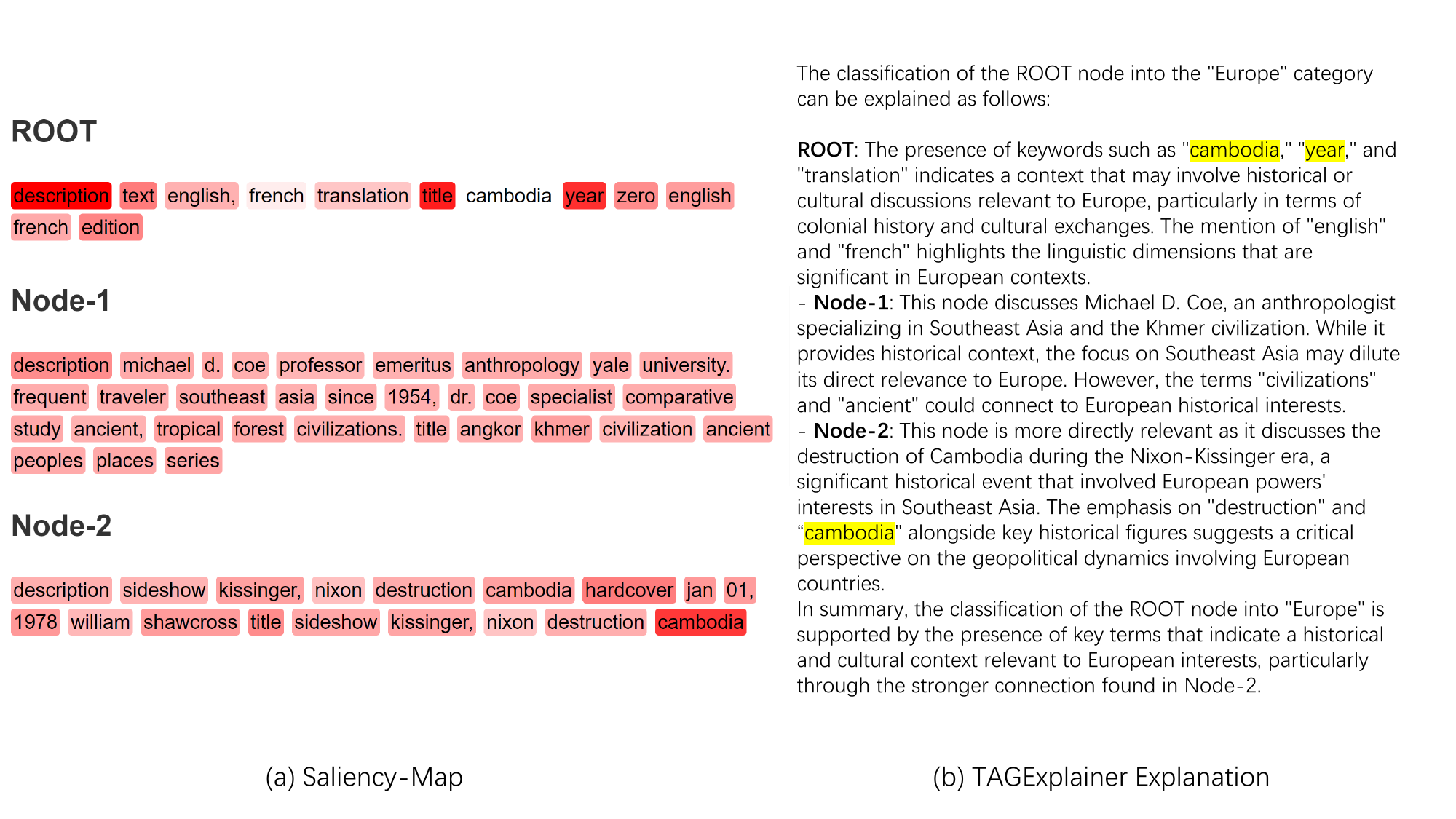}
\end{center}
\vspace{-3mm}
\caption{Example from Book-History}\label{fig:7}
\end{figure*}

We show three examples of explanations generated by GraphNarrator in Fig.~\ref{fig:visualization}, Fig.~\ref{fig:6}, Fig.~\ref{fig:7}. As depicted in Figure \ref{fig:visual-saliency-map}, we visualize the token importance of saliency-based explanation, where individual words within each node are highlighted with varying intensities of red to indicate their saliency scores. Darker red hues correspond to higher saliency scores, while lighter shades represent lower ones. In contrast, Figure \ref{fig:visual-explanation} showcases the natural language explanations generated by GraphNarrator, with key terms such as ``reinforcement learning'' and ``learning algorithm'' emphasized in yellow. This demonstrates GraphNarrator's ability to not only capture the salient information identified in the saliency map but also present it in a more accessible and interpretable manner. Compared with the saliency-based explanations that merely highlight important words, GraphNarrator goes beyond synthesizing and abstracting content across nodes. For example, in the case of Node-1.1 through Node-1.8, GraphNarrator effectively integrates the most relevant information into a coherent explanation rather than simply reproducing the input. This showcases GraphNarrator's strength in generating explanations that are more informative and contextualized than the visual saliency approach. Similar properties of GraphNarrator-generated explanations can also be seen in Fig.~\ref{fig:6}, Fig.~\ref{fig:7}.

% \FloatBarrier 

% \FloatBarrier 

% \FloatBarrier 

\section{Serialization of $\mathcal{G}$ and construction of masked token prediction task}\label{append:serialize}

\begin{figure}[h]
\begin{center}
\includegraphics[width=0.48\textwidth]{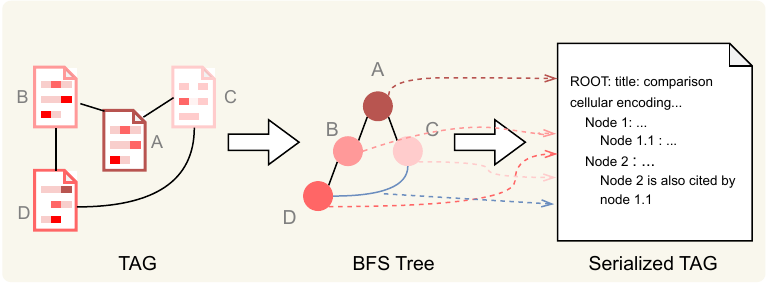}
\end{center}
\vspace{-3mm}
\caption{Illustration of text-attributed graph serialization.}\label{fig:serialization}
\end{figure}

\textbf{Serialization of $\mathcal{G}$.} To serialize the text-attributed graph $\mathcal{G}$, we use similar method as our saliency TAG verbalization method, but not include any saliency score information. Therefore, a TAG $\mathcal{G}$ is serialized by first conducting a BFS search from the root node, then using pre-order traversal to organize each node into a section in the serialized document, and adding the "cross-edges" to include the edges that connect nodes from different branches. An example is given in Fig.~\ref{fig:serialization}.

\begin{figure*}[h]
\begin{center}
\includegraphics[width=1.0\textwidth]{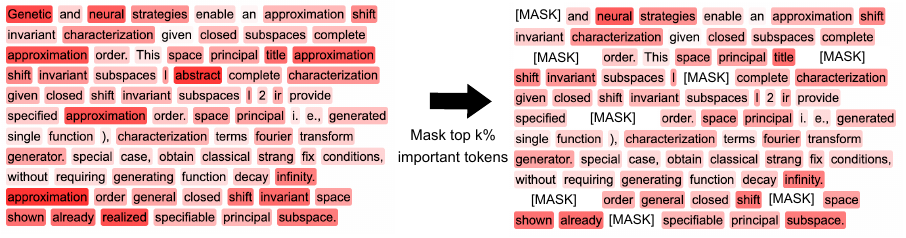}
\end{center}
\vspace{-3mm}
\caption{Illustration of masking important tokens prediction}\label{fig:masking}
\end{figure*}

\textbf{Masked token prediction with pre-trained language model:} Given a masked token prediction task, we calculate $\log P(Y|X)$, where X and Y are two arbitrary token sequences, by decomposing it into token-level predictions. Specifically, 
$$\log P(Y|X) = \sum_{i=1}^{|Y|} \log P(y_i|X, y_0, \dots, y_{i-1})$$ 
To compute $P(y_i | X, y0, \dots, y_{i-1})$, we first concatenate X with the partial sequence $[y_0, \dots, y_{i-1}]$ as input to gemma, then extract the probability of token $y_i$ from gemma's output logits.

\section{Pseudo-label quality}\label{append:pseudo-quality}
To showcase the effectiveness of our TAG Expert Iteration on improving pseudo label quality, we also conduct human evaluations on the pseudo labels used to train the Explainer LLM. The evaluation targets are same as the human evaluation on the results, stated in the experimental setting section. Results are given in Table~\ref{tab:pseudo-label-quality}. Results show that the quality of pseudo labels after the Expert Iteration is significantly better than zero-shot explanations from gpt-4o. 

\begin{table}[th]
\centering
\small
\begin{tabular}{@{}clcccc@{}}
\toprule
\multirow{2}{*}{Dataset} & \multirow{2}{*}{Method} & 
\multicolumn{4}{c}{Metrics} \\ \cmidrule(l){3-6} 
 &  &EU & DMI & SI & SeI \\ \midrule
 
\multirow{2}{*}{DBLP} 
 & GPT-4o   & 4.2 & 4.5 & 3.6 & 3.6     \\
 & Pseudo-Label & \textbf{5.2} & \textbf{5.7} & \textbf{5.7} & \textbf{4.6}   \\ \midrule

\multirow{2}{*}{Cora} 
 & GPT-4o   & 4.7 & 4.0 & 4.3 & 4.7     \\
 & Pseudo-Label  & \textbf{4.8} & \textbf{4.8 }& \textbf{5.2} & \textbf{5.0}   \\ \midrule

\multirow{2}{*}{Book-History} 
 & GPT-4o  & 4.9 & 4.6 & 3.9 & 3.9     \\
 & Pseudo-Label & \textbf{5.0} & \textbf{5.1} & \textbf{5.4} & \textbf{5.4}    \\ 
\bottomrule
\end{tabular}
\caption{Evaluation of pseudo-label quality.}
\label{tab:pseudo-label-quality}
\end{table}

\section{Additional Experiment Results}
\label{append:experiment}

\begin{figure*}[h]
\begin{center}
\includegraphics[width=1.0\textwidth]{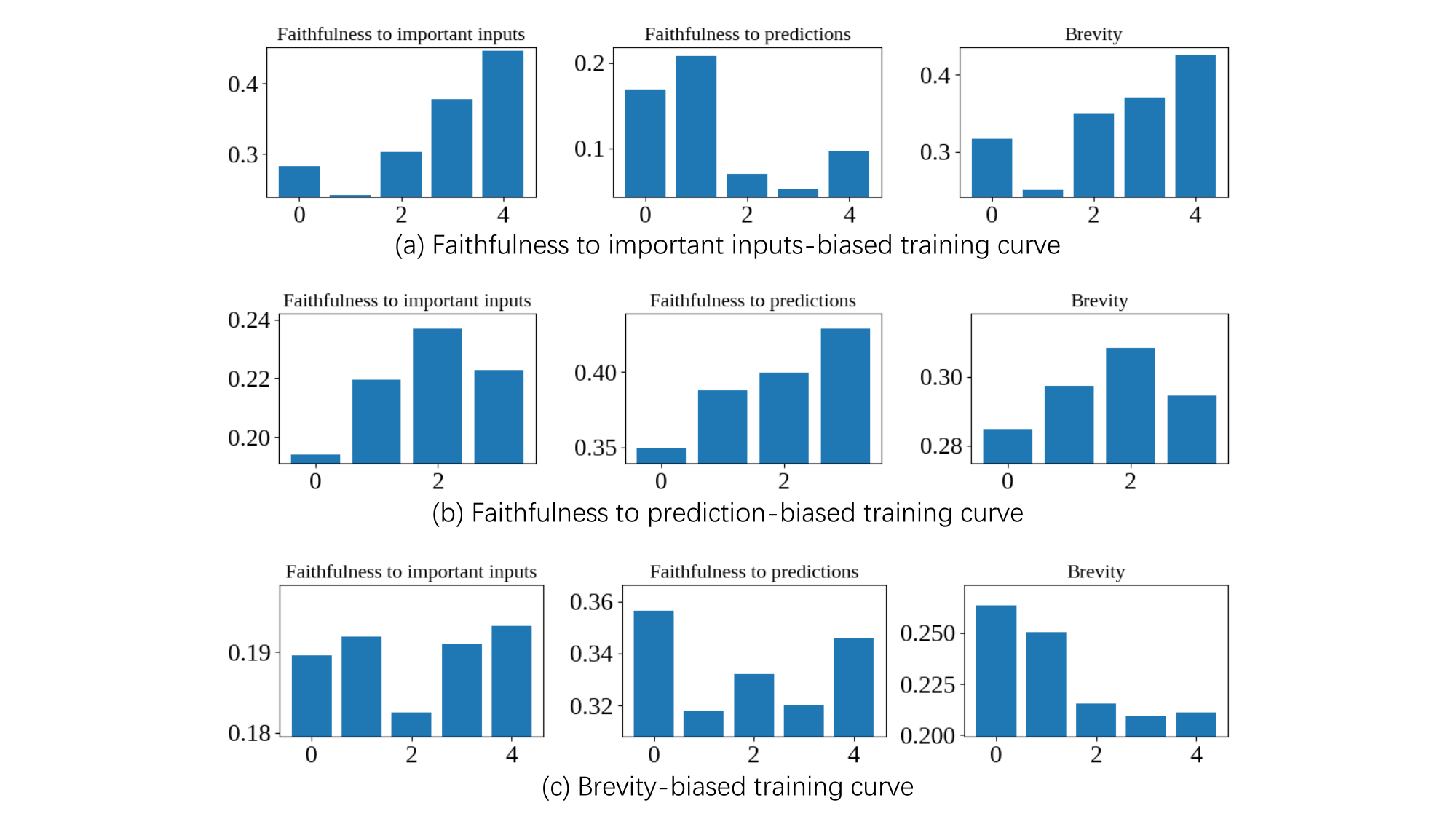}
\end{center}
\vspace{-3mm}
\caption{experiment with different customized selecting criteria}\label{fig:5}
\end{figure*}

\subsection{Experiments on different candidate selection criteria}\label{appendix:strategy}
To provide an initial validation of the effectiveness of the optimization process for the explanation generator, we conducted experiments using three extreme selection criteria: (a) a selection strategy that prioritizes only faithfulness to important inputs, (b) a strategy focusing exclusively on faithfulness to predictions, and (c) a strategy considering solely brevity. Our results (as shown in Figure \ref{fig:5}) indicate that, under each of these conditions, the corresponding metric was significantly improved. These findings suggest that the proposed framework has the capacity to selectively enhance the performance of the explanation generator with respect to specific evaluation metrics, demonstrating its adaptability and targeted optimization potential.

\subsection{The performance gain of Explainer LLM after knowledge distillation}\label{append:Distill}
\begin{table*}[h]
\centering
\small
% \resizebox{\textwidth}{!}{%
\begin{tabular}{@{}clccccc@{}}
\toprule
\multirow{2}{*}{Dataset} & \multirow{2}{*}{Method} &
\multicolumn{5}{c}{Metrics} \\ \cmidrule(l){3-7} 
 &  &  PMI-10\% ($\uparrow$) & PMI-20\% ($\uparrow$) & PMI-30\% ($\uparrow$) & Simul. ($\uparrow$) & Brevity ($\downarrow$) \\ \midrule
\multirow{2}{*}{Cora} 
 & LLaMA3.1 8B       & 0.335 & 0.278 & 0.199 & 0.78 & 0.600 \\
 & GraphNarrator    & \textbf{0.418} & \textbf{0.290} & \textbf{0.227} & \textbf{0.97} & \textbf{0.315} \\ \midrule

\multirow{2}{*}{DBLP} 
 & LLaMA3.1 8B       & 0.139 & \textbf{0.109} & 0.077 & 0.63 & 0.394 \\
 & GraphNarrator    & \textbf{0.155} & 0.108 & \textbf{0.085} & \textbf{0.95} & \textbf{0.354} \\ \midrule

\multirow{2}{*}{Book-History} 
 & LLaMA3.1 8B       & 0.465 & \textbf{0.390} & 0.281 & 0.79 & 0.735 \\
 & GraphNarrator    & \textbf{0.533} & 0.374 & \textbf{0.291} & \textbf{0.96} & \textbf{0.506} \\ 
\bottomrule
\end{tabular}%
% }
\caption{The performance of student model before (LLaMA3.1 8B) and after (our GraphNarrator) distillation. Better results are bolded.}
\label{tab:distill}
\end{table*}
We tested the performance of vanilla LLaMA3.1 8B model and the distilled version, which leads to our GraphNarrator, among three different datasets. The results demonstrated in table \ref{tab:distill} shows that the distillation process indeed promoted the quality of explanation in terms of most PMI, Simulatability and Brevity metrics.

\section{Dataset Details}\label{append:dataset}

\begin{table}[h]
\centering
\small
\begin{tabular}{c|ccc}
\toprule
             & \# Nodes & \# Edges & \# Categories \\ \hline
Cora         & 2,708    & 5,429    & 7         \\ 
DBLP         & 110,757 & 655,766 & 30        \\ 
Book-History & 41,551  & 358,574 & 12        \\ \bottomrule
\end{tabular}
\caption{Dataset Overview}
\label{tab:3}
\end{table}

We conduct experiments on 3 datasets, the basic statistics are shown in Table \ref{tab:3}. 

Cora is a network that contains computer science research papers, where each node represents a paper, and each edge represents one paper and cites the other one. Nodes in the Cora dataset are classified into seven categories: Case\_Based, Genetic\_Algorithms, Neural\_Networks, Probabilistic\_Methods, Reinforcement\_Learning, Rule\_Learning, and Theory.

DBLP dataset is a large-scale network of academic research papers, where each node represents a paper and each edge indicates a citation between two papers. Similar to the Cora dataset, which focuses on computer science research, DBLP covers a broader range of fields of study with an emphasis on computer science and related disciplines. Papers in the DBLP dataset are classified into various fields of study based on their topics. From the DBLP dataset, we extracted the top 30 most frequently occurring fields of study, along with their corresponding papers. Some of these categories include cluster analysis, cloud computing, computer science, the internet, wireless sensor networks, artificial neural networks, population, control theory, image segmentation, humanities, and image processing. These categories reflect the diverse range of research areas covered in the DBLP dataset.

Book-History dataset, extracted from the Amazon dataset~\citep{ni2019justifying}, comprises items labeled under the second-level category "History." In this dataset, each node represents a book, and edges between nodes indicate frequent co-purchases or co-views of the books. The books in the Book-History dataset are classified into 12 distinct categories: Africa, Americas, Ancient Civilizations, Arctic \& Antarctica, Asia, Australia \& Oceania, Europe, Historical Study \& Educational Resources, Middle East, Military, Russia, and World.

\section{Implementation Details}
\label{sec:impplement}

We first masked the last $5\%$ of tokens (mostly stop words and punctuations without explicit semantical contribution to the downstream tasks) based on their importance scores to form a reduced saliency-based explanation as input. We then utilized the candidate explanation generator GPT-4o-mini-2024-07-18, prompting it with a carefully designed template (see Appendix~\ref{sec:prompt} for details) and employing a one-shot learning technique to ensure consistency in the format and style of the generated explanations. For scoring and rejection sampling, we used the gemma2-2b-it model as the masked language model for the masked token prediction task to estimate the conditional probability distribution mentioned in information-theoretic objectives. In Equation~\ref{eq:f_f}, we have masked all label-related information in condition E to prevent answer leakage. During the rejection sampling phase, we found that a balanced configuration among all three objectives introduced in \ref{sec:objectives}, i.e., $\lambda_S: \lambda_F: \lambda_B = 1:1:1 $, provided stable and balanced performance across the three evaluation metrics (more customized criteria are included Appendix ~\ref{append:experiment}). We applied Expert Iteration fine-tuning by selecting 50 high-quality samples during each loop via rejection sampling. These samples were used to fine-tune the model using OpenAI API with default learning rate and batch size for 3 epochs. The final model obtained from the optimization loop served as the teacher model. We then performed knowledge distillation using the fine-tuned LLaMA-3.1-8b as the base student model, employing the LoRA technique (rank r=16 and alpha=16) for efficient fine-tuning. We minimized the cross-entropy loss between the student outputs and the teacher outputs, which resulted in our final GraphNarrator model.

For expert iteration, each iteration involves generating three scores via a Gemma-2B model deployed on a single NVIDIA H100 GPU, taking ~30 minutes per iteration. Fine-tuning and inference with GPT-4o via OpenAI API costs of ~\$5 per iteration, with training of 10, 5, and 5 iterations across the three datasets. Finally, knowledge distillation via LoRA fine-tuning of a LLaMA-3.1 8B model requires ~20 minutes per dataset on an NVIDIA A6000 GPU.

\section{Prompt Details}
\label{sec:prompt}
In our experiments, we utilize two types of prompts: one in which each token in the input is accompanied by a corresponding saliency score, and another without saliency scores. The difference between the two prompts is whether the words in the verbalized graph are accompanied by their corresponding importance scores in brackets or not.The teacher models require the inclusion of saliency scores, as they function as candidate explanation generators. The presence of saliency scores enables them to generate more accurate explanations by highlighting important tokens. In contrast, the student models do not use saliency scores; their task is to output the reasoning process of the black-box model based solely on the TAG and prediction. The student models are designed to align directly with the teacher models' outputs, ensuring consistency without requiring saliency information. All the prompts we used are given from here, due to the limited space.

\subsection{w/o Saliency Prompt}

\textbf{HumanMessage:} ``The following verbalized graph contains important words in the text of each node. These words contribute to the classification of Node 0 into one of the seven possible categories (['Case Based', 'Genetic Algorithms', 'Neural Networks', 'Probabilistic Methods', 'Reinforcement Learning', 'Rule Learning', 'Theory']). \\
Generate a concise, human-readable explanation that justifies the classification result of Node 0 by identifying and explaining the relevant inner-node features (i.e., keywords) and inter-node relationships (i.e., graph structure). The explanation should focus on how these factors contribute to the classification label.

\#\# Example

\#\#\# Verbalized Graph\\
\texttt{<verbalized-graph>}\\
ROOT: title experiments real time decision algorithms abstract real time decision algorithms class incremental resource bounded horvitz, 89 anytime dean, 93 algorithms evaluating influence diagrams. present test domain real time decision algorithms, results experiments several real time decision algorithms domain. results demonstrate high performance two algorithms, decision evaluation variant incremental probabilisitic inference dambrosio, 93 variant algorithm suggested goldszmidt, goldszmidt, 95 ], pk reduced. discuss implications experimental results explore broader applicability algorithms. \\
Node-1: title learning policies partially observable environments scaling abstract partially observable markov decision processes pomdp model decision problems agent tries maximize reward face limited noisy sensor feedback. \\
Node-1.1: title formal framework speedup learning problems solutions abstract speedup learning seeks improve computational efficiency problem solving experience. paper, develop formal framework learning efficient problem solving random problems solutions.  \\
Node-1.2: title acting uncertainty discrete bayesian models mobile robot navigation abstract discrete bayesian models used model uncertainty mobile robot navigation, question actions chosen remains largely unexplored. \\
Node-1.3: title incremental methods computing bounds partially observable markov decision processes abstract partially observable markov decision processes pomdps allow one model complex dynamic decision control problems include action outcome uncertainty imperfect observabil ity.  \\
Node-1.4: title learning sorting decision trees pomdps abstract pomdps general models sequential decisions actions observations probabilistic. many problems interest formulated pomdps, yet use pomdps limited lack effective algorithms. recently started change number problems robot navigation planning beginning formulated solved pomdps.
``Node-1.5: title approximating optimal policies partially observable stochastic domains abstract problem making optimal decisions uncertain conditions central artificial intelligence. state world known times, world modeled markov decision process mdp ).\\
Node-1.6: title efficient dynamic programming updates partially observable markov decision processes abstract examine problem performing exact dynamic programming updates partially observable markov decision processes pomdps computational complexity viewpoint.  \\
Node-2: title efficient inference bayes networks combinatorial optimization problem abstract number exact algorithms developed perform probabilistic inference bayesian belief networks recent years. \\
Node-2.1: title sensitivities alternative conditional probabilities bayesian belief networks abstract show alternative way representing bayesian belief network sensitivities probability distributions. representation equivalent traditional representation conditional probabilities, makes dependencies nodes apparent intuitively easy understand.  \\
Node-2.2: title algebraic techniques efficient inference bayesian networks abstract number exact algorithms developed perform probabilistic inference bayesian belief networks recent years. algorithms use graph theoretic techniques analyze exploit network topology. \\
Node-2.3: title interpretation complex scenes using bayesian networks abstract object recognition systems, interactions objects scene ignored best interpretation considered set hypothesized objects matches greatest number image features.  \\
Node-2.4: title case based probability factoring bayesian belief networks abstract bayesian network inference formulated combinatorial optimization problem, concerning computation optimal factoring distribution represented net. since determination optimal factoring computationally hard problem, heuristic greedy strategies able find approximations optimal factoring usually adopted. present paper investigate alternative approach based combination genetic algorithms ga case based reasoning cbr ).\\
\texttt{</verbalized-graph>}''
``
\#\#\# Classification Label
Probabilistic Methods

\#\#\# Reasoning

0. Graph Structure Reconstruction:

In the provided verbalized graph, The ROOT node (first line) is the target for classification. \\
Single-digit indexed nodes are direct neighbors of ROOT. \\
Double-digit indexed nodes are: \\
  - Two hops away from ROOT \\
  - Direct children of their parent node \\
More digits indexed nodes follow the same principle as described above. \\

Thus, the graph structure of this verbalized graph is: 
\begin{itemize}
    \item ROOT
    \begin{itemize}
        \item Node-1
        \begin{itemize}
            \item Node-1.1
            \item Node-1.2
            \item Node-1.3
            \item Node-1.4
            \item Node-1.5
            \item Node-1.6
        \end{itemize}
        \item Node-2
        \begin{itemize}
            \item Node-2.1
            \item Node-2.2
            \item Node-2.3
            \item Node-2.4
        \end{itemize}
    \end{itemize}
\end{itemize}

1. Word-Level Evaluation: \\
Detect important terms for the classification label. \\
Quantitatively, the importance (saliency) scores behind each word in the verbalized graph are calculated by a post-hoc XAI (explainable AI) algorithm. \\
Semantically, keywords synonymous with or closely related to label, 'Probabilistic Methods' (such as 'probabilistic', 'probability', and 'inference'), are semantically influential in determining the classification because they directly indicate the node's relevance to the 'Probabilistic Methods' category. \\
2. Graph-Level Aggregation: \\
Node 1 does not contain as many directly relevant keywords. Yet, it still discusses POMDPs, which are probabilistic models, indicating that it contributes to the classification in a less significant way. \\
Direct neighbors like Node 2, which contains the keyword 'probabilistic', have a substantial influence on ROOT node classification due to their close proximity to the ROOT node. \\
Indirect neighbors, Nodes-2.1 to Node-2.4, connected through Node 2, contain keywords like 'probabilistic' and 'probability', which further reinforce the classification of ROOT. \\
Meanwhile, the indirect neighbors connected through Node 1, Node-1.1 to Node-1.6, are less relevant because they do not include important keywords directly related to 'Probabilistic Methods'. \\

\#\#\# Free-Text Explanation
```markdown
The classification of ROOT node into the "Probabilistic Methods" category can be explained as follows:

ROOT: The presence of keywords like "probabilistic" and "inference" within ROOT node suggests that the paper is focused on topics central to probabilistic methods, such as performing probabilistic inference and evaluating influence diagrams. \\
  - Node-1: While Node 1 focuses on POMDPs, which are also probabilistic models, it is slightly less directly related to the probabilistic inference methods emphasized in ROOT node. \\
    - Node-1.1 ~ Node-1.6: they are less relevant to probabilistic methods and thus contribute less to the classification. \\
  - Node-2: This node discusses "probabilistic inference" and "Bayesian networks," both of which are foundational to probabilistic methods. The strong thematic connection between ROOT node and Node 2 reinforces the classification. \\
    - Node-2.1 ~ Node-2.4: they also discuss probabilistic topics, further supporting the classification. \\

In summary, the classification of ROOT node into "Probabilistic Methods" is well-supported by the strong presence of key probabilistic terms within the node itself and its direct connection to Node 2, which also focuses on probabilistic inference. Indirect neighbors through Node 2 further reinforce this classification.''

``\#\# Task

\#\#\# Verbalized Graph \\
\texttt{<verbalized-graph>\{document\}</verbalized-graph>}

\#\#\# Classification Label \\
\texttt{\{label\}}

\#\#\# Reasoning

\#\#\# Free-Text Explanation

\textit{(P.S.: 1. make sure to complete both the reasoning section and then Free-Text Explanation section with the same structure as exemplified above. \\
2. make good use of the importance (saliency) score behind each word as your guidance to generate the better explanation. However, it is not necessary to directly quote the saliency score. \\
3. use the \textbf{whole} graph structure you constructed during reasoning for the format of the explanation. Indents and node indexes are necessary, which represent the hierarchy of the graph.)}

\subsection{w/ Saliency Prompt}

\textbf{HumanMessage:} ``The following verbalized graph contains important words in the text of each nodes. These words (each with corresponding importance score in the bracket) contributes to the classification of Node 0 into one of the seven possible categories (['Case Based', 'Genetic Algorithms', 'Neural Networks', 'Probabilistic Methods', 'Reinforcement Learning', 'Rule Learning', 'Theory']). \\
Generate a concise, human-readable explanation that justifies the classification result of Node 0 by identifying and explaining the relevant inner-node features (i.e., keywords) and inter-node relationships (i.e., graph structure). The explanation should focus on how these factors contribute to the classification label.

\#\# Example

\#\#\# Verbalized Graph\\
\texttt{<verbalized-graph>}\\
ROOT: title(9.13) experiments(7.56) real(2.52) time(2.41) decision(5.20) algorithms(7.18) abstract(12.01) real(3.17) time(2.82) decision(5.46) algorithms(10.39) class(4.34) incremental(2.60) resource(4.50) bounded(5.79) horvitz,(2.67) 89(4.58) anytime(6.66) dean,(4.92) 93(5.03) algorithms(7.94) evaluating(4.75) influence(7.70) diagrams. \\
Node-1: title(12.47) learning(12.87) policies(9.77) partially(3.11) observable(2.82) environments(5.58) scaling(9.39) abstract(10.80) partially(4.42) observable(2.62) markov(4.50) decision(5.75) processes(4.53) pomdp(9.69) model(11.47) decision(7.63) problems(7.18) agent(12.00) tries(3.13) maximize(3.05) reward(6.03) face(2.13) limited(2.17) noisy(8.96) sensor(6.27) feedback. ''
``Node-1.1: title(0.95) formal(0.36) framework(0.41) speedup(0.35) learning(0.41) problems(0.48) solutions(0.48) abstract(1.14) speedup(0.33) learning(0.61) seeks(0.57) improve(0.27) computational(0.50) efficiency(0.35) problem(0.37) solving(0.41) experience.(0.57) paper,(0.70) develop(0.53) formal(0.40) framework(0.38) learning(0.37) efficient(0.40) problem(0.34) solving(0.32) random(0.54) problems(0.32) solutions. \\
Node-1.2: title(2.30) acting(0.98) uncertainty(2.31) discrete(1.03) bayesian(1.13) models(0.94) mobile(1.03) robot(1.75) navigation(1.12) abstract(2.80) discrete(1.18) bayesian(0.97) models(0.81) used(0.66) model(0.56) uncertainty(1.79) mobile(0.77) robot(1.55) navigation,(0.66) question(0.64) actions(1.17) chosen(0.67) remains(0.72) largely(0.56) unexplored. \\
Node-1.3: title(1.50) incremental(0.64) methods(0.50) computing(0.59) bounds(1.09) partially(0.24) observable(0.21) markov(0.31) decision(0.64) processes(0.38) abstract(0.97) partially(0.25) observable(0.21) markov(0.32) decision(0.58) processes(0.36) pomdps(0.21) allow(0.54) one(0.36) model(0.47) complex(0.38) dynamic(0.60) decision(0.76) control(0.38) problems(0.53) include(0.31) action(0.82) outcome(0.55) uncertainty(0.61) imperfect(0.54) observabil(0.22) ity.(0.36)  \\
Node-1.4: title(0.98) learning(1.07) sorting(1.63) decision(1.56) trees(2.00) pomdps(1.04) abstract(1.34) pomdps(1.10) general(0.42) models(0.59) sequential(0.99) decisions(0.93) actions(0.63) observations(1.27) probabilistic.(0.59) many(0.37) problems(1.06) interest(0.66) formulated(0.98) pomdps. ''

``Node-1.5: title(0.90) approximating(0.42) optimal(0.69) policies(0.65) partially(1.13) observable(0.41) stochastic(0.50) domains(0.63) abstract(1.25) problem(0.51) making(0.22) optimal(0.40) decisions(0.42) uncertain(1.01) conditions(0.80) central(0.56) artificial(0.82) intelligence. \\
Node-1.6: title(1.58) efficient(1.08) dynamic(0.83) programming(1.15) updates(2.24) partially(0.70) observable(0.55) markov(0.87) decision(1.19) processes(0.86) abstract(1.67) examine(0.99) problem(0.72) performing(0.50) exact(0.70) dynamic(0.57) programming(0.75) updates(1.48) partially(1.26) observable(0.58) markov(0.78) decision(1.58) processes(0.78) pomdps(1.18) computational(1.04) complexity(0.75) viewpoint. \\
Node-2: title(14.46) efficient(7.56) inference(7.77) bayes(5.83) networks(10.92) combinatorial(4.43) optimization(7.56) problem(7.08) abstract(20.68) number(4.43) exact(10.23) algorithms(14.38) developed(3.37) perform(4.58) probabilistic(5.11) inference(6.22) bayesian(9.36) belief(43.68) networks(17.76) recent(7.91) years. \\
Node-2.1: title(0.96) sensitivities(0.32) alternative(0.73) conditional(0.51) probabilities(0.29) bayesian(0.53) belief(1.27) networks(0.93) abstract(1.29) show(0.82) alternative(0.44) way(0.32) representing(0.60) bayesian(0.48) belief(1.94) network(1.51) sensitivities(0.25) probability(0.69) distributions.''

``Node-2.2: title(1.81) algebraic(1.21) techniques(0.51) efficient(0.73) inference(0.70) bayesian(0.55) networks(1.04) abstract(1.57) number(0.37) exact(0.77) algorithms(1.75) developed(0.34) perform(0.43) probabilistic(0.35) inference(0.59) bayesian(0.50) belief(1.91) networks(1.06) recent(1.20) years. \\
Node-2.3: title(1.43) interpretation(0.76) complex(0.42) scenes(0.80) using(0.46) bayesian(0.87) networks(0.89) abstract(1.12) object(0.85) recognition(1.42) systems,(0.46) interactions(0.45) objects(0.43) scene(0.56) ignored(0.38) best(0.26) interpretation(0.69) considered(0.24) set(0.22) hypothesized(0.20) objects(0.39) matches(0.23) greatest(0.26) number(0.20) image(0.50) features. \\
Node-2.4: title(1.04) case(0.45) based(0.40) probability(0.95) factoring(0.36) bayesian(0.53) belief(1.33) networks(1.03) abstract(1.04) bayesian(0.90) network(0.97) inference(0.87) formulated(0.55) combinatorial(0.30) optimization(0.73) problem,(0.41) concerning(0.65) computation(0.55) optimal(0.54) factoring(0.40) distribution(1.54) represented(0.70) net.\\
\texttt{</verbalized-graph>}

\#\#\# Classification Label
Probabilistic Methods

\#\#\# Reasoning

0. Graph Structure Reconstruction:

In the provided verbalized graph, The ROOT node (first line) is the target for classification. \\
Single-digit indexed nodes are direct neighbors of ROOT. \\
Double-digit indexed nodes are: \\
  - Two hops away from ROOT \\
  - Direct children of their parent node \\
More digits indexed nodes follow the same principle as described above.''

``Thus, the graph structure of this verbalized graph is: 
\begin{itemize}
    \item ROOT
    \begin{itemize}
        \item Node-1
        \begin{itemize}
            \item Node-1.1
            \item Node-1.2
            \item Node-1.3
            \item Node-1.4
            \item Node-1.5
            \item Node-1.6
        \end{itemize}
        \item Node-2
        \begin{itemize}
            \item Node-2.1
            \item Node-2.2
            \item Node-2.3
            \item Node-2.4
        \end{itemize}
    \end{itemize}
\end{itemize}

1. Word-Level Evaluation: \\

Detect important terms for the classification label. \\
Quantitatively, the importance (saliency) scores behind each word in the verbalized graph are calculated by a post-hoc XAI (explainable AI) algorithm. \\
Semantically, keywords synonymous with or closely related to label, 'Probabilistic Methods' (such as 'probabilistic', 'probability', and 'inference'), are semantically influential in determining the classification because they directly indicate the node's relevance to the 'Probabilistic Methods' category. \\
2. Graph-Level Aggregation: \\

Node 1 does not contain as many directly relevant keywords. Yet, it still discusses POMDPs, which are probabilistic models, indicating that it contributes to the classification in a less significant way. \\
Direct neighbors like Node 2, which contains the keyword 'probabilistic', have a substantial influence on ROOT node classification due to their close proximity to the ROOT node. \\
Indirect neighbors, Nodes-2.1 to Node-2.4, connected through Node 2, contain keywords like 'probabilistic' and 'probability', which further reinforce the classification of ROOT. \\
Meanwhile, the indirect neighbors connected through Node 1, Node-1.1 to Node-1.6, are less relevant because they do not include important keywords directly related to 'Probabilistic Methods'. \\

\#\#\# Free-Text Explanation
```markdown
The classification of ROOT node into the "Probabilistic Methods" category can be explained as follows:

ROOT: The presence of keywords like "probabilistic" and "inference" within ROOT node suggests that the paper is focused on topics central to probabilistic methods, such as performing probabilistic inference and evaluating influence diagrams. \\
  - Node-1: While Node 1 focuses on POMDPs, which are also probabilistic models, it is slightly less directly related to the probabilistic inference methods emphasized in ROOT node. \\
    - Node-1.1 ~ Node-1.6: they are less relevant to probabilistic methods and thus contribute less to the classification. \\
  - Node-2: This node discusses "probabilistic inference" and "Bayesian networks," both of which are foundational to probabilistic methods. The strong thematic connection between ROOT node and Node 2 reinforces the classification. \\
    - Node-2.1 ~ Node-2.4: they also discuss probabilistic topics, further supporting the classification. \\

In summary, the classification of ROOT node into "Probabilistic Methods" is well-supported by the strong presence of key probabilistic terms within the node itself and its direct connection to Node 2, which also focuses on probabilistic inference. Indirect neighbors through Node 2 further reinforce this classification.''

``\#\# Task

\#\#\# Verbalized Graph \\
\texttt{<verbalized-graph>\{document\}</verbalized-graph>}

\#\#\# Classification Label \\
\texttt{\{label\}}

\#\#\# Reasoning

\#\#\# Free-Text Explanation

\textit{(P.S.: 1. make sure to complete both the reasoning section and then Free-Text Explanation section with the same structure as exemplified above. \\
2. make good use of the importance (saliency) score behind each word as your guidance to generate the better explanation. However, it is not necessary to directly quote the saliency score. \\
3. use the \textbf{whole} graph structure you constructed during reasoning for the format of the explanation. Indents and node indexes are necessary, which represent the hierarchy of the graph.)}''

\end{document}